\documentclass[10pt,twocolumn,letterpaper]{article}

\usepackage{iccv}
\usepackage{times}
\usepackage{epsfig}
\usepackage{graphicx}
\usepackage{amsmath}
\usepackage{amssymb}

\usepackage{booktabs}
\usepackage[ruled,vlined]{algorithm2e}
\usepackage{color}
\usepackage{comment}
\usepackage{enumitem}
\usepackage[export]{adjustbox}
\usepackage{tabularx}
\usepackage[dvipsnames]{xcolor}
\usepackage{subfig}
\usepackage{multirow}
\usepackage{booktabs}
\usepackage{xr}
\usepackage{pifont}
\usepackage[hyphens]{url}
\usepackage{sidecap}
\usepackage{tablefootnote}

\usepackage[font=small,labelfont=bf,skip=2pt]{caption}

\usepackage[pagebackref=true,breaklinks=true,letterpaper=true,colorlinks,bookmarks=false]{hyperref}

\usepackage[capitalize]{cleveref}
\crefname{section}{Sec.}{Secs.}
\Crefname{section}{Section}{Sections}
\Crefname{table}{Table}{Tables}
\crefname{table}{Tab.}{Tabs.}

\newcommand{\Paragraph}[1]{\vspace{0mm} \noindent \textbf{#1} \hspace{0mm}}

\newcommand{\cmark}{\ding{51}}%
\newcommand{\xmark}{\ding{55}}%

\newcommand{\tablestyle}[2]{\setlength{\tabcolsep}{#1}\renewcommand{\arraystretch}{#2}\centering\footnotesize}

\iccvfinalcopy 

\def\iccvPaperID{3061} 

\newcommand{\revise}[1]{\textcolor{black}{#1}}

\ificcvfinal\fi

\begin{document}

\title{MPT: Mesh Pre-Training with Transformers for Human Pose and Mesh Reconstruction}

\author{Kevin Lin, Chung-Ching Lin, Lin Liang, Zicheng Liu, Lijuan Wang\\
Microsoft\\
{\tt\small \{keli, chungching.lin, lliang, zliu, lijuanw\}@microsoft.com}
}

\maketitle
\ificcvfinal\thispagestyle{empty}\fi

\begin{abstract}

Traditional methods of reconstructing 3D human pose and mesh from single images rely on paired image-mesh datasets, which can be difficult and expensive to obtain. Due to this limitation, model scalability is constrained as well as reconstruction performance. Towards addressing the challenge, we introduce Mesh Pre-Training (MPT), an effective pre-training strategy that leverages large amounts of MoCap data to effectively perform pre-training at scale. We introduce the use of MoCap-generated heatmaps as input representations to the mesh regression transformer and propose a Masked Heatmap Modeling approach for improving pre-training performance. This study demonstrates that pre-training using the proposed MPT allows our models to perform effective inference without requiring fine-tuning. We further show that fine-tuning the pre-trained MPT model considerably improves the accuracy of human mesh reconstruction from single images. Experimental results show that MPT outperforms previous state-of-the-art methods on Human3.6M and 3DPW datasets. As a further application, we benchmark and study MPT on the task of 3D hand reconstruction, showing that our generic pre-training scheme generalizes well to hand pose estimation and achieves promising reconstruction performance.

\end{abstract}

\section{Introduction}\label{sec:intro}

\begin{figure}[ht]
\begin{center}
\includegraphics[trim=0 0 0 0, clip,width=1.0\columnwidth]{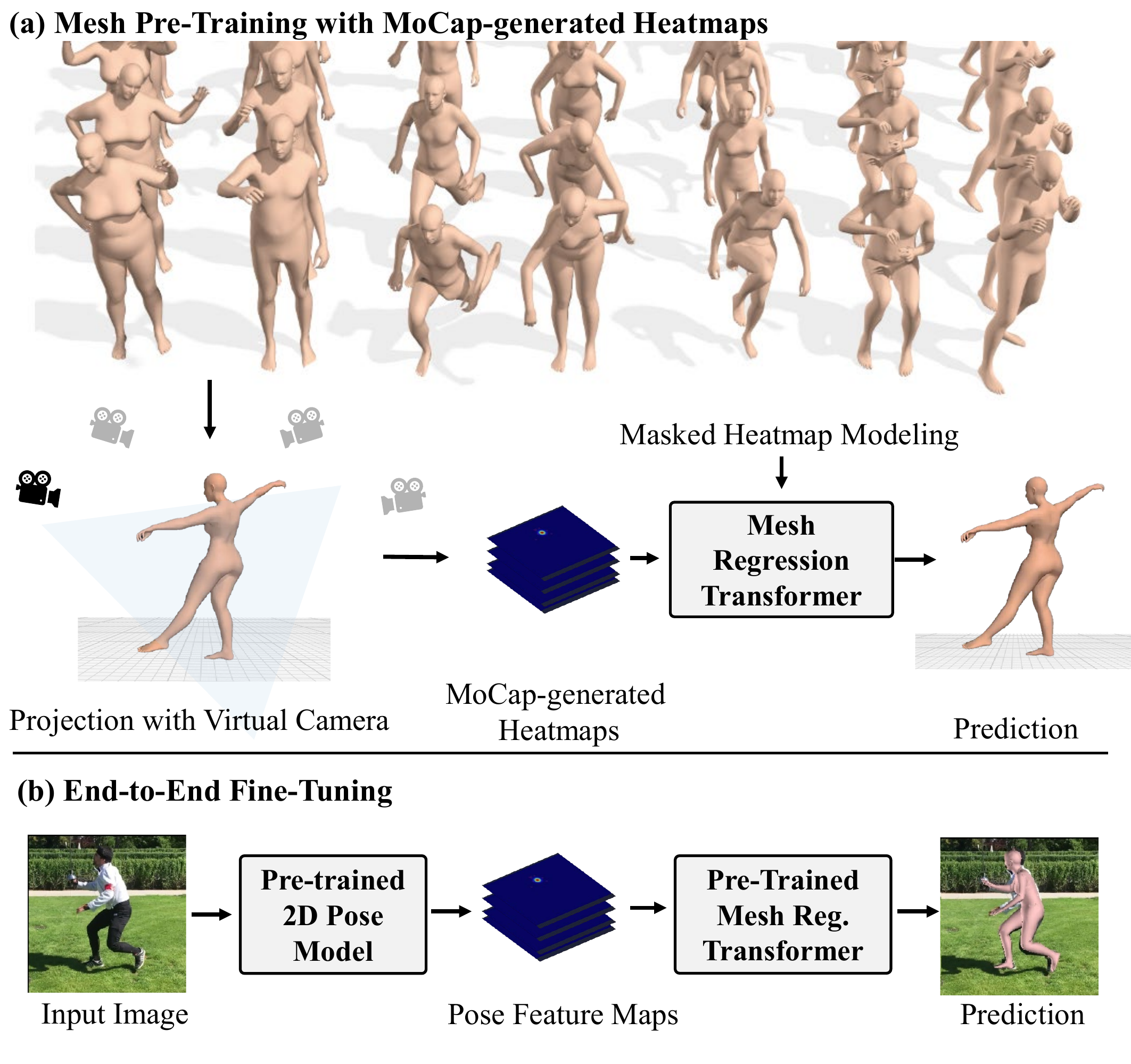}
\caption{
Summary of our pretrain-finetune strategy. (a) In our proposed Mesh Pre-Training, we pre-train mesh regression transformer using MoCap-generated heatmaps to learn human pose and shape knowledge. (b) \revise{After pre-training, we use an off-the-shelf 2D pose model to extract pose feature maps, which are then fed to the mesh regression transformer. Through end-to-end fine-tuning, our model learns to reconstruct 3D human pose and mesh from the input image.}
} 
\vspace{-6mm}
\label{fig:fig1}
\end{center}
\end{figure}

3D human pose and mesh reconstruction from a single image is a challenging task in computer vision~\cite{kanazawa2018end,kolotouros2019learning,kocabas2021pare,kolotouros2019convolutional,Moon_2020_ECCV_I2L-MeshNet,Choi_2020_ECCV_Pose2Mesh,cho2022FastMETRO,li2022cliff}, which involves estimating the 3D coordinates of human body joints and mesh vertices from a 2D image. The task has various applications in areas such as human motion analysis and human-centric event understanding.  
Recent advances in transformer models~\cite{lin2020end,lin2021mesh,cho2022FastMETRO} have shown remarkable success in this task. Most of the models are, however, trained in a supervised manner using paired image-mesh datasets, which are costly to acquire in practice. This has limited model scalability and performance, and also the development of 3D pose reconstruction.

To address the challenge, prior works~\cite{ martinez2017simple,Zheng_2021_ICCV,ci2019optimizing,pavlakos2018learning,Choi_2020_ECCV_Pose2Mesh} attempted to use 3D mesh data, such as motion capture (MoCap) data~\cite{cmu-mocap,mahmood2019amass}, for training a 2D-to-3D lifting model that projects the 2D joint coordinates into 3D space. Such 3D mesh data has proven to be beneficial in learning detailed skeleton articulations. 
Despite providing accurate 3D joint coordinates along with the body meshes, MoCap data generally lack corresponding RGB images. The majority of these approaches, therefore, do not involve the use of images in learning and could be prone to depth ambiguity.

In this paper, we propose an effective pre-training method called Mesh Pre-Training (MPT) that leverages large amounts of MoCap data and learns with MoCap-generated heatmaps. Instead of directly performing a 2D-to-3D lifting task, as shown in Figure~\ref{fig:fig1}(a), we propose to pre-train the mesh regression transformer by using MoCap-generated heatmaps, which are synthesized from the 2D joint coordinates obtained from virtual cameras for each 3D mesh sample. These heatmaps serve as input representations for the mesh regression transformer during pre-training. To facilitate self-attention learning with such input representations, we propose a Masked Heatmap Modeling (MHM) approach to improve pre-training performance.

After pre-training, as shown in Figure~\ref{fig:fig1}(b), we use an off-the-shelf 2D human pose estimation model to extract pose feature maps, which are then fed to the mesh regression transformer for human mesh reconstruction. \revise{We then fine-tune both the 2D human pose estimation model and the mesh regression transformer in an end-to-end manner. As we will show in the experiments, our model learns to extract pose feature maps with a more general focus on multiple body joints. Accordingly, the pose feature maps together with the pre-trained mesh regression transformer contribute to the reconstruction of high-fidelity human meshes.}

Our MPT model is pre-trained on a large-scale MoCap dataset consisting of 2 million human meshes. After pre-training, we fine-tune and evaluate the model on target datasets for single-image 3D human pose and mesh reconstruction. Experimental results demonstrate that our proposed MPT method advances the state-of-the-art performance on multiple public benchmarks, including the Human3.6M and 3DPW datasets. In addition, our experimental results suggests that the proposed MPT enables inference capability without the need of fine-tuning.

Furthermore, we demonstrate the versatility of our approach by applying MPT to the task of single-image 3D hand reconstruction, achieving state-of-the-art results on the FreiHAND dataset. The proposed MPT approach provides a promising direction for scaling up pre-training for single-image 3D human pose and mesh reconstruction without the need for paired image-mesh datasets.

In summary, we make the following contributions.
\begin{itemize}
\item{We propose an effective pre-training method, called Mesh Pre-Training (MPT), for the 3D human pose and mesh reconstruction from single images.}
\item{We introduce the use of MoCap-generated heatmaps with Masked Heatmap Modeling (MHM) for pre-training mesh regression transformer.}

\item{The proposed method achieves new state-of-the-art performance on multiple benchmarks including Human3.6M, 3DPW, and FreiHAND.}

\end{itemize}

\section{Related Works}\label{sec:related}

\Paragraph{Single-image human pose and mesh reconstruction:} Prior works can be clustered into two categories: parametric and non-parametric approaches. Parametric approaches~\cite{kolotouros2019learning,kanazawa2018end,tung2017self,guan2009estimating,kocabas2019vibe,lassner2017unite,pavlakos2018learning} typically adopt the SMPL model~\cite{loper2015smpl} and regress SMPL parameters to generate human meshes. While SMPL model has shown great success and is convenient and robust to pose variations, it is challenging to estimate accurate SMPL parameters from a single image. Recent studies~\cite{pavlakos2020human,moon2020pose2pose,zhang2020learning, tung2017self, guler2018densepose} have been focusing on various auxiliary supervisions such as improving 2D re-projection~\cite{li2022cliff} or extending it to video-based methods~\cite{kocabas2019vibe,wan2021encoder,choi2021beyond} to improve the estimation of SMPL parameters.  

Different from adopting SMPL as the regression target, non-parametric approaches~\cite{Choi_2020_ECCV_Pose2Mesh, Moon_2020_ECCV_I2L-MeshNet, kolotouros2019convolutional, lin2020end, lin2021mesh, cho2022FastMETRO} aim to predict the 3D coordinates of body joints and mesh vertices directly from the input image. Researchers have explored graph convolutional neural networks~\cite{Choi_2020_ECCV_Pose2Mesh,kolotouros2019convolutional} as well as transformer architecture~\cite{lin2020end, lin2021mesh, cho2022FastMETRO} which is effective in modeling vertex-vertex and vertex-joint interactions for improving the reconstruction performance. 

Our work draws inspiration from Pose2Mesh~\cite{Choi_2020_ECCV_Pose2Mesh}, a related study that employs a multi-stage 2D-to-3D lifting pipeline. Pose2Mesh begins by detecting 2D body joint locations and then elevating these coordinates into 3D space. It subsequently reconstructs a 3D mesh based on the lifted 3D joint coordinates. 
In comparison, we propose to use MoCap-generated heatmaps to pre-train the mesh regression transformer. Our design enables end-to-end training and considerably improves the performance of human mesh reconstruction. 

\begin{figure*}[t]
\begin{center}
\includegraphics[trim=0 0 0 0, clip,width=1.0\textwidth]{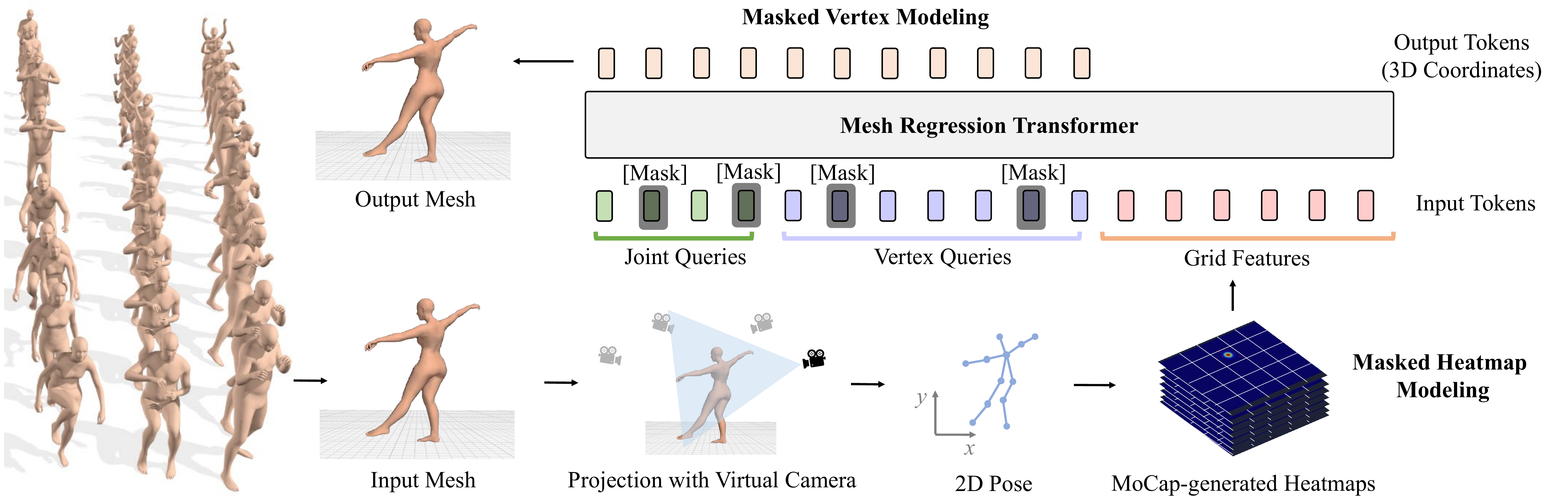}
\vspace{-4mm}
\caption{
\textbf{Overview of the proposed Mesh Pre-Training (MPT).} Given a human mesh which is sampled from MoCap data, we project its 3D human pose to 2D by randomly selecting a virtual camera. We then synthesize the heatmaps to represent the projected 2D human pose. Our mesh regression transformer takes three types of input tokens, including joint queries, vertex queries, and MoCap-generated heatmaps. We then pre-train the transformer to predict the 3D coordinates of body joints and mesh vertices given the input tokens. During pre-training, we perform Masked Heatmap Modeling and Masked Vertex Modeling to improve the robustness of the mesh regression transformer.
} 
\vspace{-6mm}
\label{fig:overview}
\end{center}
\end{figure*}

\Paragraph{Synthetic training data generation:} Previous studies have explored 3D graphics rendering~\cite{rogez2016mocap,varol17_surreal,baradel2021leveraging, cai2021playing} to generate large amount of image-mesh pairs. However, models trained on synthetic data may not perform well on real images due to the domain gap between synthetic and real data~\cite{ebadi2022psp,tobin2017domain,tremblay2018training}. Recent studies~\cite{li2022cliff,Moon_2022_CVPRW_NeuralAnnot,joo2021exemplar} have proposed training a human mesh annotator to generate 3D pseudo labels on real images. Unlike existing works that collect image-mesh pairs, our method uses MoCap-generated heatmaps to pre-train models using only 3D mesh data. Our method does not rely on the photorealism of the synthetic data generation, and provides an effective way to leverage 3D mesh datasets that lack associated RGB images.

\Paragraph{Human motion capture (MoCap) datasets:} There are many optical marker-based motion capture data available~\cite{akhter2015pose,mandery2015kit,muller2009efficient,troje2002decomposing}, including CMU~\cite{cmu-mocap}, and AMASS~\cite{mahmood2019amass}. MoCap data records a variety of human body movements, which is useful for human motion generation and analysis~\cite{kocabas2019vibe,petrovich2022temos,tevet2022motionclip,tevet2022human}. Since MoCap data is usually captured by the infrared (IR) sensors and performers have to wear special cloth and markers, there are usually no RGB images available. To deal with the problem, prior works~\cite{rogez2016mocap,varol17_surreal,baradel2021leveraging} proposed to leverage 3D graphics engines to synthesize RGB images for the collection of image-mesh pairs. More recently, researchers have proposed to train a motion discriminator~\cite{kocabas2019vibe} to help improve video-based human mesh reconstruction. Different from prior works, we use MoCap datasets to pre-train our mesh regression transformer which improves the accuracy of 3D pose and mesh reconstruction from a single image.

\section{Method}\label{sec:method}

Our approach uses a two-stage training scheme that consists of (\textit{i}) mesh pre-training, (\textit{ii}) end-to-end fine-tuning. Figure~\ref{fig:fig1} illustrates our approach using an example. First, in the pre-training stage (Figure~\ref{fig:fig1}(a)), we pre-train the mesh regression transformer to learn human mesh reconstruction by using MoCap-generated heatmaps. These heatmaps are synthesized from 2D joint coordinates which are obtained by projecting 3D joints with a randomly selected virtual camera. Second, \revise{Figure~\ref{fig:fig1}(b) illustrates our framework at the fine-tuning stage. We use an off-the-shelf pre-trained 2D pose estimation model to extract pose feature maps, which are then fed into the mesh regression transformer for human mesh reconstruction. We then perform end-to-end fine-tuning on the target dataset.}

\subsection{Pre-Training and MoCap-generated Heatmaps}

As shown in Figure~\ref{fig:overview}, our mesh regression transformer takes three types of tokens as inputs, including joint queries, vertex queries, and MoCap-generated heatmaps. The mesh regression transformer is asked to directly regress the 3D coordinates of body joints and mesh vertices from MoCap-generated heatmaps.

We pre-train our mesh regression transformer using a large scale MoCap dataset (\textit{e.g.}, AMASS dataset~\cite{mahmood2019amass}). Unlike existing works that rely on image-mesh pairs, the proposed Mesh Pre-Training (MPT) is conducted by using 3D mesh data without RGB images. As AMASS dataset consists of sequences of SMPL human meshes, we sparsely sample the human meshes from each motion capture sequence, and then obtain the corresponding 3D human poses using SMPL regressor. This results in a total of 2 million meshes. Note that all the meshes are normalized and centered at the origin. We pre-define 4 virtual cameras above the head positions of the human meshes. For each mesh and each virtual camera, we obtain the projected 2D joint coordinates which are then used to synthesize the joint heatmaps. In this way, we generate 8 million heatmap-mesh pairs which are used in pre-training. More details on the heatmap generation are discussed next.

\subsubsection{Synthesizing Heatmaps}

Given a mesh, we first project the 3D pose to 2D by randomly selecting a virtual camera. We then represent the 2D pose in the form of heatmaps~\cite{cheng2020higherhrnet,sun2019deep,WangSCJDZLMTWLX19} (also called confidence maps~\cite{cao2017realtime}). To be specific, we generate a set of heatmaps $\mathbf{S}$ based on the projected 2D pose. The set $\mathbf{S} = (\mathbf{S}_1, \mathbf{S}_2, \dots, \mathbf{S}_K)$ has $K$ heatmaps, one per joint, where $K$ is the total number of body joints, $\mathbf{S}_j \in \mathbb{R}^{w\times h}$ for $j\in \{1,2,\dots,K\}$. We set $K=17$ in our experiments following the body joint definition in COCO dataset~\cite{lin2014microsoft}. Each heatmap depicts the 2D position of a specified body joint.  Let $x_j$ be the 2D coordinate of the body joint $j$. The value at location $\mathbf{p}$ in the heatmap $\mathbf{S}_{j}$ is \begin{equation}\label{eqn:confidencemap}{\mathbf{S}_j}(\mathbf{p}) = \exp \left(-\frac{ \left| \left| \mathbf{p} - x_j \right| \right|^2_2 }{\sigma^2} \right),
\end{equation}where $\sigma=3$ following the literature~\cite{sun2019deep}.

In order to input the heatmaps to the transformer model, we concatenate the heatmaps $\mathbf{S}$ along the channel dimension to form a 3D tensor, which is of size $(W\times H \times C)$, and $W=H=224$. Similar to ViT~\cite{dosovitskiy2020image}, we split the tensor into non-overlapping patches using a patch partition module~\cite{dosovitskiy2020image,liu2021Swin}. Each patch is then treated as an input token (or grid feature~\cite{jiang2020defense}) to the transformer model. In our implementation, we use a patch size of $(8 \times 8)$. That is, the feature dimension of each patch is $(8 \times 8 \times C)$. Finally, we apply a multi-layer perceptron (MLP) layer to make the dimension of grid features consistent with the hidden size of the transformer model.

\subsubsection{Masked Heatmap Modeling}
To facilitate the learning of transformer self-attention with MoCap-generated heatmaps, during pre-training, we randomly mask some of the joints in the MoCap-generated heatmaps, and ask the mesh regression transformer to predict all the 3D body joints and mesh vertices. Unlike existing works~\cite{lin2020end,lin2021mesh} that only used Masked Vertex Modeling to randomly mask out some of the query tokens, we perform masking on the heatmaps. Our masking mechanism is in spirit similar to simulating the real heatmaps obtained from the off-the-shelf 2D pose estimation model, where some joints might be missing. In addition to the masking, we also apply data augmentation to the heatmaps, including adding Gaussian noise and joint coordinate jittering.

\subsubsection{Pre-Training Objective}

Our pre-training objective is a regression task conditioned on the MoCap-generated heatmaps.  
To regress the mesh vertices, following the literature~\cite{kolotouros2019convolutional,lin2020end,lin2021mesh,cho2022FastMETRO}, we use $L_{1}$ loss to minimize the differences between the predicted vertices $V_{3D}$ and the ground truth vertices $\bar{V}_{3D}$:\begin{equation}\begin{aligned}%
\label{eqn:vertex-loss}%
\mathcal{L}_{V} = \frac{1}{M}\sum_{i=1}^{M} \left| \left| V_{3D}-\bar{V}_{3D} \right| \right|_1,
\end{aligned}
\end{equation}
where $\bar{V}_{3D} \in \mathbb{R}^{M \times 3}$, and $M$ is the total number of vertices.

In addition, we minimize the differences between the predicted joints $J_{3D}$ and the ground truth joints $\bar{J}_{3D}$:
\begin{equation}\begin{aligned}%
\label{eqn:3dpose-loss}%
\mathcal{L}_{J} = \frac{1}{K}\sum_{i=1}^{K} \left| \left| J_{3D}-\bar{J}_{3D} \right| \right|_1,\end{aligned}\end{equation} where $K$ is the total number of 3D joints.

Note that the 3D joints can be regressed from the mesh vertices using a pre-defined matrix $G$~\cite{Choi_2020_ECCV_Pose2Mesh,kanazawa2018end,kolotouros2019convolutional,kolotouros2019learning,li2022cliff,cho2022FastMETRO}. We also apply supervision on the regressed 3D joints:
\begin{equation}\begin{aligned}%
\label{eqn:3dpose-loss-reg}%
\mathcal{L}_{J}^{reg} = \frac{1}{K}\sum_{i=1}^{K} \left| \left| J_{3D}^{reg}-\bar{J}_{3D} \right| \right|_1,
\end{aligned}
\end{equation}
where $J_{3D}^{reg} = GV_{3D}$, and $G \in \mathbb{R}^{K \times M}$.

Following the common practice~\cite{kanazawa2018end, kolotouros2019convolutional,kolotouros2019learning,cho2022FastMETRO,li2022cliff,lin2021mesh,lin2020end}, we also employ the 2D re-projection loss. Given the predicted 3D joints, we project the predicted 3D joints to 2D using the estimated camera parameters. We then minimize the errors between the projected 2D joints $J_{2D}$ and the ground truth 2D joints $\bar{J}_{2D}$:\begin{equation} \begin{aligned}%
\label{eqn:2dpose-loss}%
\mathcal{L}_{J}^{proj} = \frac{1}{K}\sum_{i=1}^{K} \left| \left| J_{2D}-\bar{J}_{2D} \right| \right|_1.
\end{aligned}
\end{equation}

Finally, our pre-training objective can be written as
\begin{equation}\begin{aligned}%
\label{eqn:all-loss}%
\mathcal{L}_{\text{pre-train}} = \mathcal{L}_{V} + \mathcal{L}_{J} + \mathcal{L}_{J}^{reg} + \mathcal{L}_{J}^{proj}.
\end{aligned}
\end{equation}

\begin{figure*}
\begin{center}
\includegraphics[trim=0 0 0 0, clip,width=0.195\textwidth]{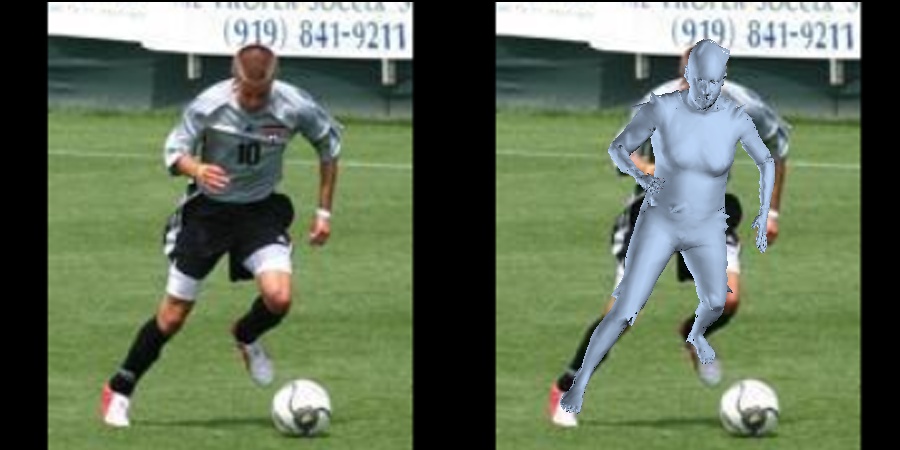}
\includegraphics[trim=0 0 0 0, clip,width=0.195\textwidth]{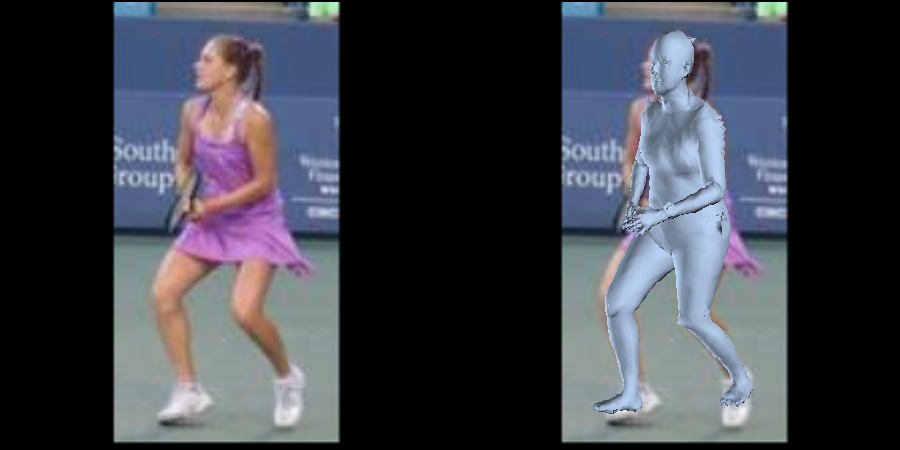}
\includegraphics[trim=0 0 0 0, clip,width=0.195\textwidth]{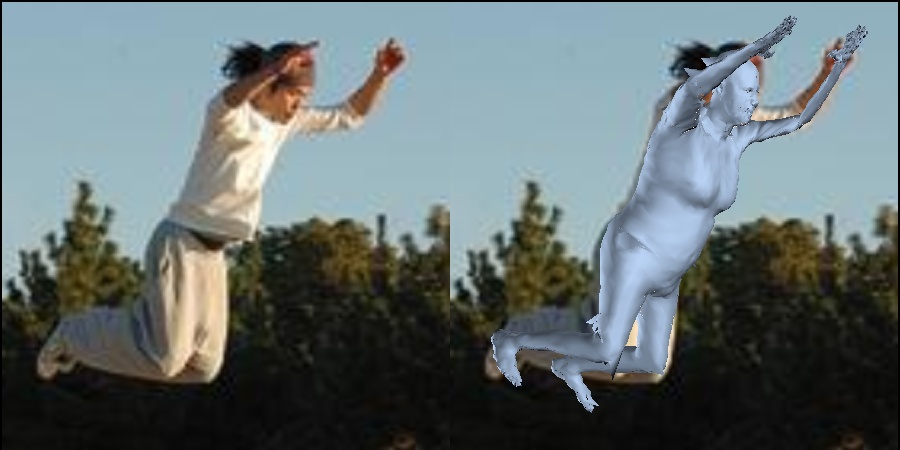}
\includegraphics[trim=0 0 0 0, clip,width=0.195\textwidth]{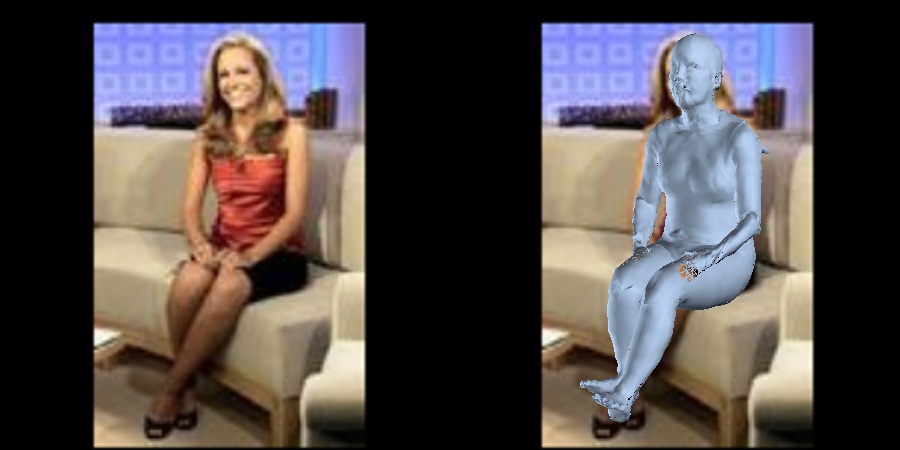}
\includegraphics[trim=0 0 0 0, clip,width=0.195\textwidth]{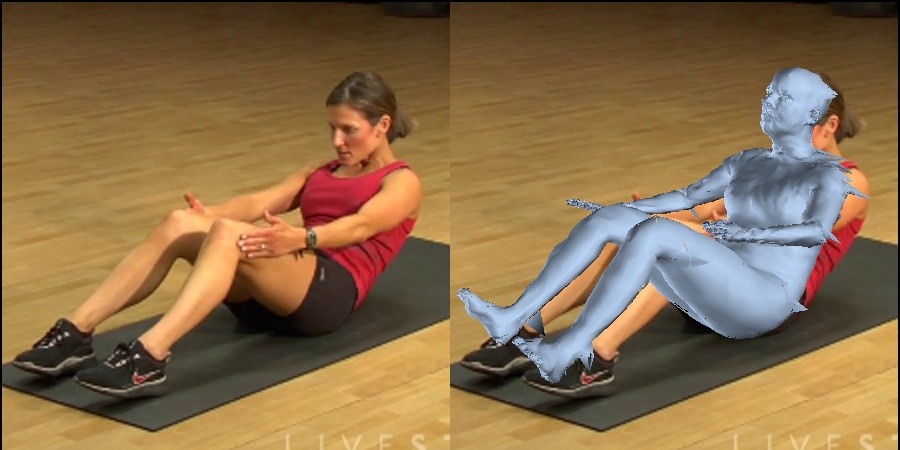}\\
\includegraphics[trim=0 0 0 0, clip,width=0.195\textwidth]{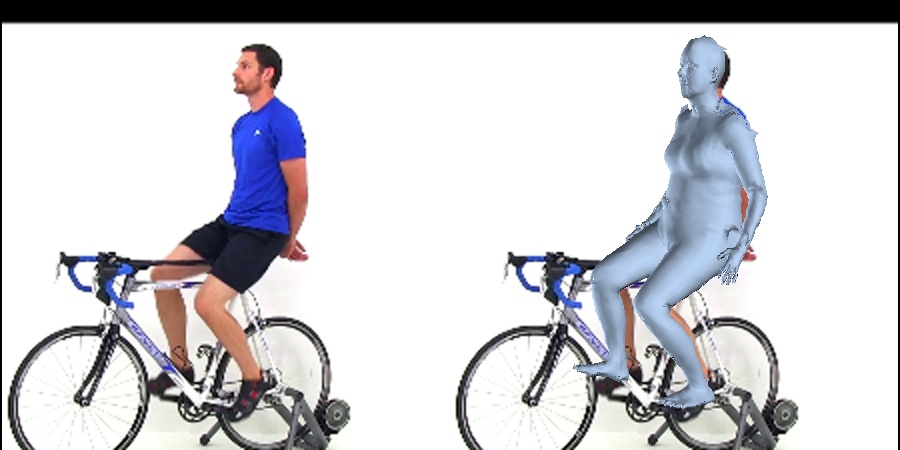}
\includegraphics[trim=0 0 0 0, clip,width=0.195\textwidth]{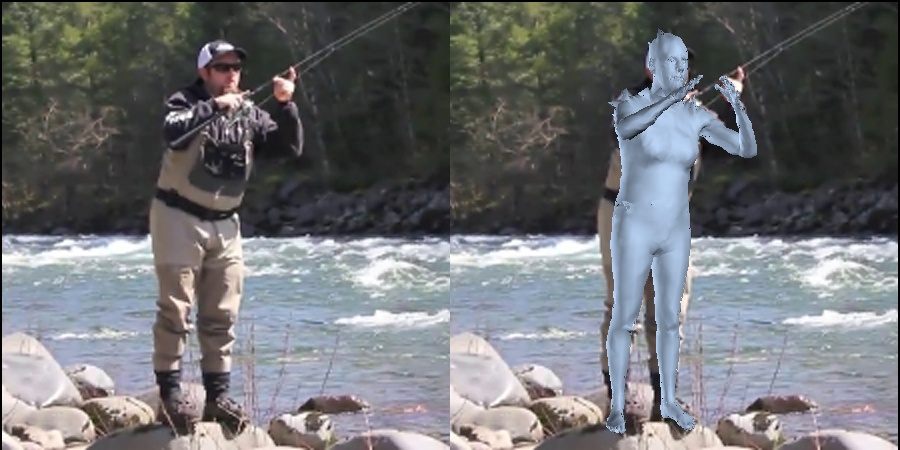}
\includegraphics[trim=0 0 0 0, clip,width=0.195\textwidth]{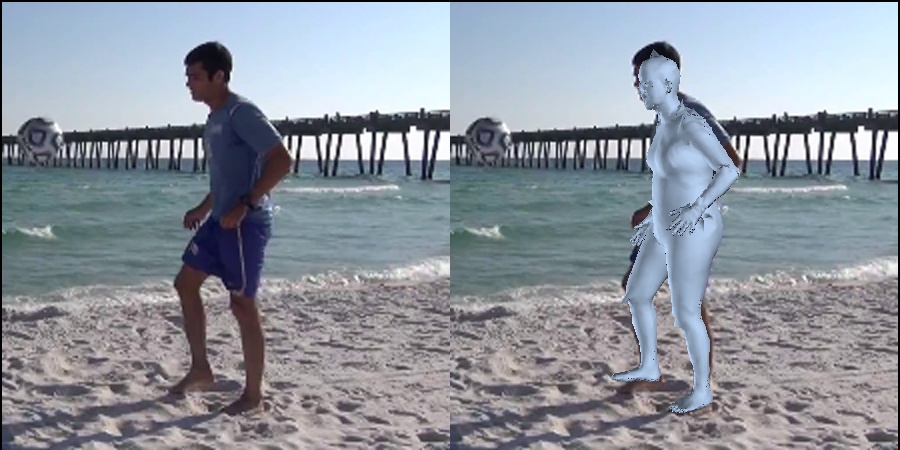}
\includegraphics[trim=0 0 0 0, clip,width=0.195\textwidth]{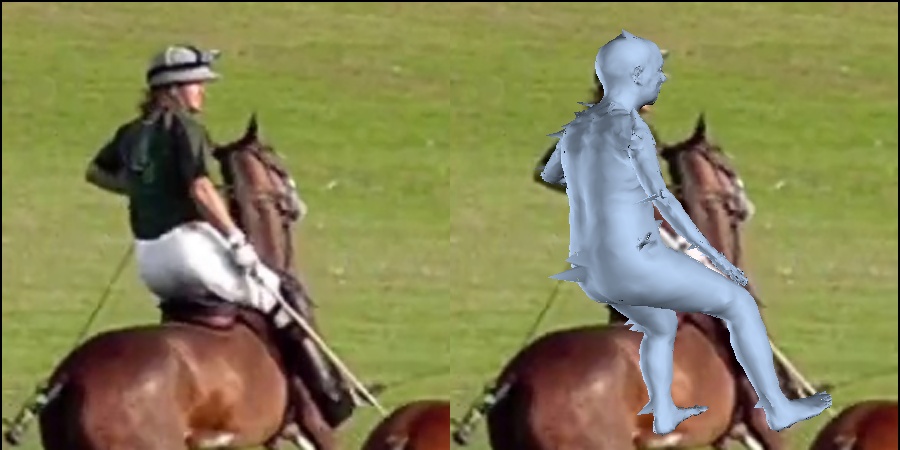}
\includegraphics[trim=0 0 0 0, clip,width=0.195\textwidth]{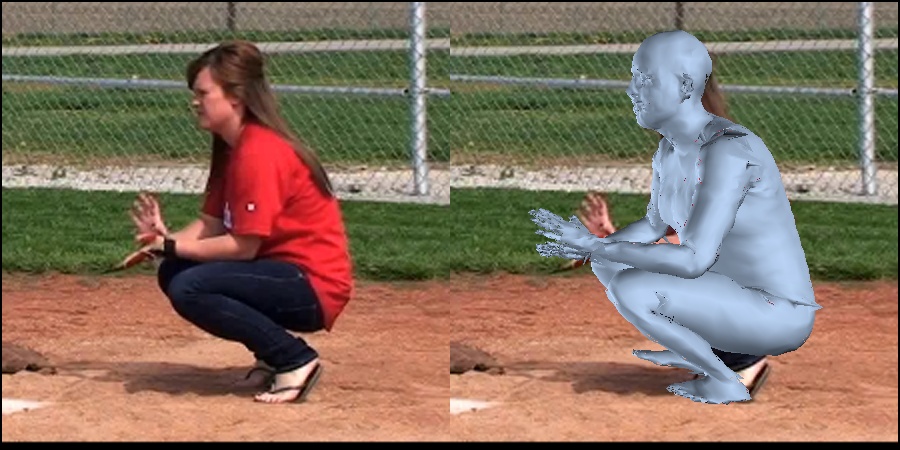}\\
\includegraphics[trim=0 0 0 0, clip,width=0.195\textwidth]{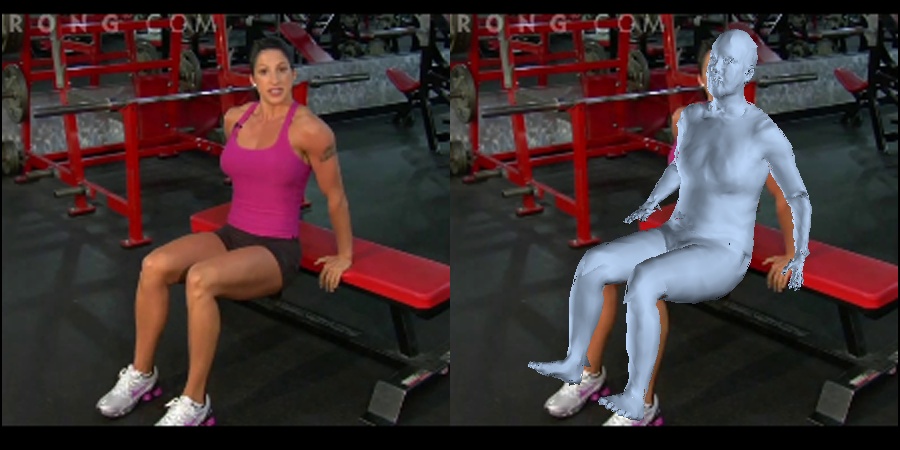}
\includegraphics[trim=0 0 0 0, clip,width=0.195\textwidth]{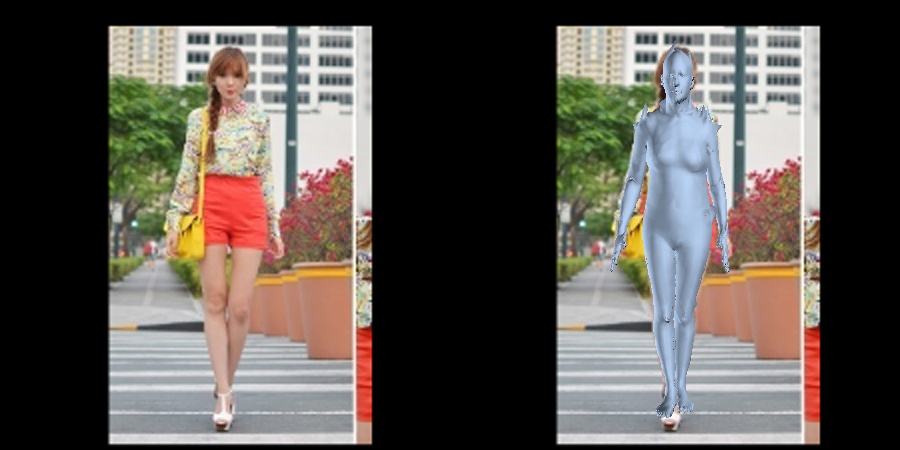}
\includegraphics[trim=0 0 0 0, clip,width=0.195\textwidth]{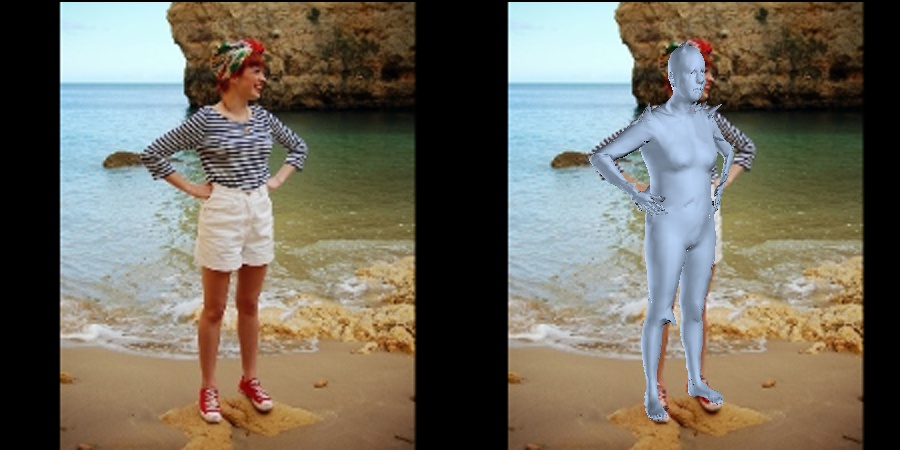}
\includegraphics[trim=0 0 0 0, clip,width=0.195\textwidth]{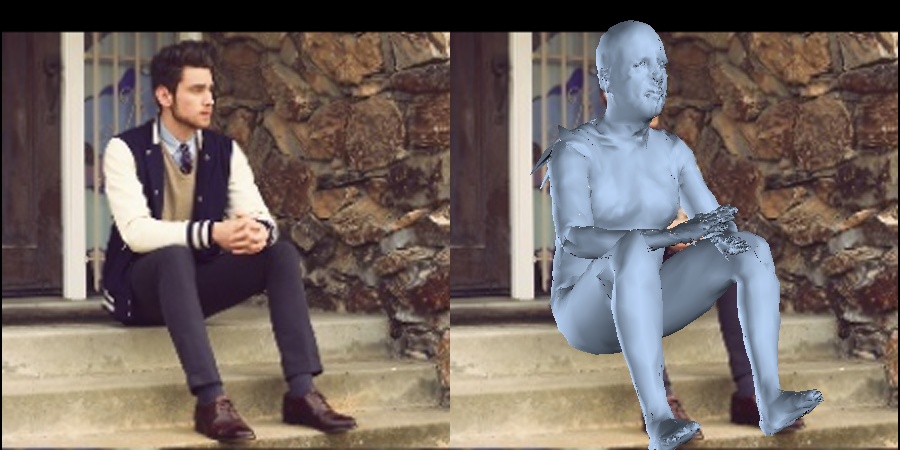}
\includegraphics[trim=0 0 0 0, clip,width=0.195\textwidth]{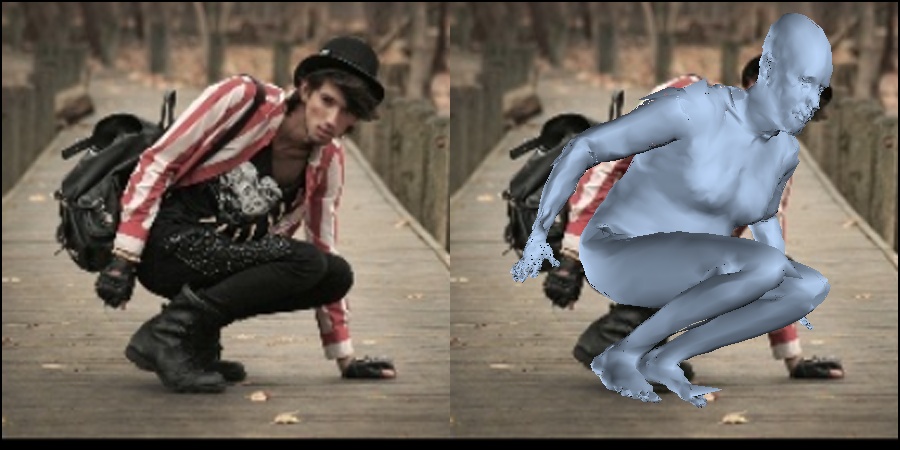}
\caption{
Testing without fine-tuning. After pre-training, we test our MPT model on real images directly. MPT generates reasonable human meshes on real images without any fine-tuning. This demonstrates the generalizability of our MPT model. 
} 
\label{fig:zero-shot}
\vspace{-4mm}
\end{center}
\end{figure*} 

\subsection{Testing without Fine-tuning}

\revise{After pre-training, the pre-trained MPT model can be directly applied to any real images for human mesh reconstruction. 
Given an input image, we use an off-the-shelf 2D pose estimation model to extract the pose feature maps. A 2D pose estimation model typically predicts a set of heatmaps (one heatmap per joint), followed by a series of post-processing to obtain the 2D joint coordinates. We remove those post-processing operations. The remaining network is used as our backbone network to extract pose feature maps from a given image.}

\revise{We then input the extracted feature maps to the mesh regression transformer for 3D human pose and mesh reconstruction. In this way, our pre-trained MPT model is capable of human mesh reconstruction in a plug-and-play fashion. }

It is worth noting that there are many well-performing pre-trained 2D pose estimation models~\cite{sun2019deep,cao2017realtime,cheng2020higherhrnet,WangSCJDZLMTWLX19,newell2016stacked,chen2018cascaded,xiao2018simple} available, and we use a recently developed one (\textit{e.g.,} HigherHRNet~\cite{cheng2020higherhrnet}) in our main experiments. 
In addition, we will show that our proposed MPT is not sensitive to the choices of 2D pose estimation model.

\subsection{End-to-End Fine-Tuning}

After pre-training, the pre-trained MPT model can be fine-tuned to the target dataset for more accurate reconstruction. Given the image-mesh pairs from the target dataset, we fine-tune both the backbone network and mesh regression transformer in an end-to-end manner. We use the same loss function as that in our pre-training stage. 

Similar to the existing studies~\cite{kanazawa2018end, kolotouros2019convolutional,kolotouros2019learning,Moon_2020_ECCV_I2L-MeshNet,cho2022FastMETRO,li2022cliff,lin2021mesh,lin2020end}, we fine-tune our model using a mixture of 2D and 3D datasets. We calculate the losses as long as the ground truths are available.

\subsection{Implementation Details}
Our mesh regression transformer is in spirit similar to METRO~\cite{lin2020end}. It consists of multiple transformer layers 
to regress the 3D coordinates of mesh vertices and body joints. An important difference of our model is that we take \revise{pose feature maps} as inputs to the transformer network. \revise{We empirically observed the design is effective to bridge the 2D pose estimation network and the pre-trained mesh regression transformer.}
\section{Experimental Results}\label{sec:exp}

In this section, we first discuss the datasets we used in pre-training and fine-tuning. We then present the performance comparison with existing state-of-the-arts on public benchmarks. Finally, we provide a detailed ablation study to verify the effectiveness of the proposed training scheme.

\subsection{Datasets and Evaluation Metrics}

We conduct pre-training on AMASS collection~\cite{mahmood2019amass}, which consists of $24$ MoCap datasets. AMASS provides a unified human pose and mesh representations based on SMPL~\cite{loper2015smpl}. Each sequence records a motion movement of a subject. In total, there are $500$ subjects with $17,916$ motions. The total length of all the sequences is about $3,772$ minutes. In total, there are about $25,088,088$ frames, and each frame has a human mesh. To avoid the redundancy between the neighbor frames, we sparsely sample 2 million meshes from AMASS. After projection using 4 virtual cameras, we obtain 8 million heatmap-mesh pairs for pre-training.

We fine-tune our model using a mixture of 2D and 3D datasets, including Human3.6M~\cite{ionescu2014human3}, MuCo-3DHP~\cite{mehta2018single}, UP-3D~\cite{lassner2017unite},  COCO~\cite{lin2014microsoft}, and MPII~\cite{andriluka14cvpr}. Note that these datasets are commonly used in literature~\cite{Choi_2020_ECCV_Pose2Mesh,lin2020end,lin2021mesh,cho2022FastMETRO}. After that, we evaluate our model on Human3.6M using P2 protocol~\cite{kanazawa2018end,kolotouros2019learning}. When conducting experiments on 3DPW~\cite{vonMarcard2018}, we follow the prior works~\cite{kocabas2019vibe,lin2020end,lin2021mesh,cho2022FastMETRO} and fine-tune with 3DPW training data. We then evaluate the results on 3DPW test set.

Following literature~\cite{kanazawa2018end, kolotouros2019convolutional,kolotouros2019learning,cho2022FastMETRO,li2022cliff}, we use three standard metrics for evaluation, including Mean Per Joint Position Error (MPJPE)~\cite{ionescu2014human3}, Procrustes Analysis with MPJPE (PA-MPJPE)~\cite{zhou2018monocap}, and Mean Per Vertex Error (MPVE)~\cite{pavlakos2018learning}. The unit of the metrics is millimeter (mm).

\begin{table}[t]
\centering
\tablestyle{0.8pt}{1.1} 
\begin{tabular}{lcccccc}
    \toprule[1.5pt]
    \multirow{1}{*}{} & \multicolumn{3}{c}{3DPW} & & \multicolumn{2}{c}{Human3.6M}\\ 
    \cline{2-4}\cline{6-7}
	Method  & MPVE $\downarrow$ & MPJPE $\downarrow$ & PA-MPJPE $\downarrow$ & &  MPJPE $\downarrow$ & PA-MPJPE $\downarrow$ \\
	\midrule
	HMR~\cite{kanazawa2018end} & $-$ & $-$ & 81.3 && 88.0 & 56.8  \\
	GraphCMR~\cite{kolotouros2019convolutional} & $-$ & $-$ & 70.2 && $-$ & 50.1\\
	SPIN~\cite{kolotouros2019learning} & 116.4 & 96.9 & 59.2 && 62.5 & 41.1\\
	Pose2Mesh~\cite{Choi_2020_ECCV_Pose2Mesh} & $-$ & 89.2 & 58.9 && 64.9 & 47.0\\
	I2LMeshNet~\cite{Moon_2020_ECCV_I2L-MeshNet} & $-$ & 93.2 & 57.7 && 55.7 & 41.1\\
	PyMAF~\cite{zhang2021pymaf} & 110.1 & 92.8 & 58.9 && 57.7 & 40.5\\
	ROMP~\cite{sun2021monocular} & 93.4 & 76.7 & 47.3 && $-$ & $-$\\
	VIBE~\cite{kocabas2019vibe} & 99.1 & 82.0 & 51.9 && 65.6 & 41.4\\
    METRO~\cite{lin2020end} & 88.2 & 77.1 & 47.9 && 54.0 & 36.7\\
    THUNDER~\cite{zanfir2021thundr} & 88.0 & 74.8 & 51.5 && 48.0 & 34.9\\
    PARE~\cite{kocabas2021pare} & 88.6 & 74.5 & 46.5 && $-$ & $-$\\
    Graphormer~\cite{lin2021mesh} & 87.7 & 74.7 & 45.6 && 51.2 & 34.5\\
	FastMETRO~\cite{cho2022FastMETRO} & 84.1 & 73.5 & 44.6 && 52.2 & 33.7\\
	CLIFF~\cite{li2022cliff} & 81.2 & 69.0 & 43.0 && 47.1 & 32.7\\
    \midrule
    MPT (Ours) & $\textbf{79.4}$ & $\textbf{65.9}$ & $\textbf{42.8}$ && $\textbf{45.3}$ & $\textbf{31.7}$\\
	\bottomrule[1.5pt]
\end{tabular}
\caption{Performance comparison with the previous state-of-the-art methods on 3DPW and Human3.6M datasets.}
\label{tbl:compare-h36m-3dpw}
\vspace{-4mm}
\end{table}

\subsection{Main Results}

We compare our method with the existing state-of-the-art approaches on Human3.6M~\cite{ionescu2014human3} and 3DPW~\cite{vonMarcard2018} datasets. We present our pretrain-then-finetune results in Table~\ref{tbl:compare-h36m-3dpw}. Our method outperforms the previous works on both datasets, including the recent transformer-based methods~\cite{lin2020end,lin2021mesh,zanfir2021thundr,cho2022FastMETRO}. 

It is worth noting that, CLIFF~\cite{li2022cliff} was the state-of-the-art approach on the two datasets. CLIFF additionally leverages the camera focal length and bounding box information to calculate 2D re-projection loss on the non-cropped input image. In contrast, we do not have such post-processing, and still achieve better results, especially on MPJPE.

\subsection{Analysis}

\Paragraph{Effectiveness of Mesh Pre-Training:} We study whether our pre-training is useful for performance improvements. We conduct experiments on Human3.6M, and Table~\ref{tbl:pretrain-analysis} shows the comparison with different training configurations including with or without mesh pre-training, and different datasets for fine-tuning. Because our model architecture is similar to METRO~\cite{lin2020end}, we include it as the reference. We also include the well-known HMR~\cite{kanazawa2018end} as reference.

In Table~\ref{tbl:pretrain-analysis}, our MPT improves the performance across different training configurations considered. When fine-tuning our pre-trained model using Human3.6M only, as shown in the sixth row, MPT achieves $35.5$ PA-MPJPE, which is better than $39.2$ PA-MPJPE of our non-pretrain model. We observe similar findings when using multiple datasets for fine-tuning. Our MPT achieves $31.5$ PA-MPJPE, and is better than $32.4$ PA-MPJPE of our non-pretrain model.

\Paragraph{Testing without Fine-Tuning:} Given the pre-trained MPT model, we directly evaluate MPT model on real images without any fine-tuning. As shown in the fifth row of Table~\ref{tbl:pretrain-analysis}, we obtain a performance of $58.4$ PA-MPJPE. Although it lags behind the supervised fully fine-tuned models, the result is comparable to the well-known HMR~\cite{kanazawa2018end}. Figure~\ref{fig:zero-shot} shows the qualitative results. We see that MPT is capable of generating human meshes on real images without the need of fine-tuning.

\begin{table}[t]
\tablestyle{1pt}{1.1} 
\centering
\begin{tabular}{lcccc}
    \toprule[1.5pt]
	Method  & MPT & FT & MPJPE $\downarrow$ & PA-MPJPE $\downarrow$ \\
	\midrule
	HMR~\cite{kanazawa2018end} & \xmark & Mixed Datasets & 88.0 & 56.8\\
	METRO~\cite{lin2020end} & \xmark & Mixed Datasets & 54.0 & 36.7\\
	\midrule
	MPT  & \xmark & Human3.6M & 59.1 & 39.2\\
	MPT  & \xmark & Mixed Datasets & 46.6 & 32.4\\
	\midrule
	MPT (Test w/o fine-tune) & \cmark & \xmark & 88.9 & 58.4\\
	MPT  & \cmark & Human3.6M & 53.3 & 35.5\\
	MPT  & \cmark & Mixed Datasets & 45.3 & 31.7\\
	\bottomrule[1.5pt]
\end{tabular}
\caption{Pre-training analysis. We conduct training with different configurations, and then evaluate results on Human3.6M validation set. MPT: Mesh Pre-Training. FT: Fine-Tuning.}
\label{tbl:pretrain-analysis}
\vspace{-4mm}
\end{table}

\begin{figure}[t]
\vspace{-2mm}
\centering
 \subfloat[(a)][Learning Convergence\label{fig:converge}]{
   \includegraphics[trim=0 0 0 0, clip,width=0.48\linewidth]{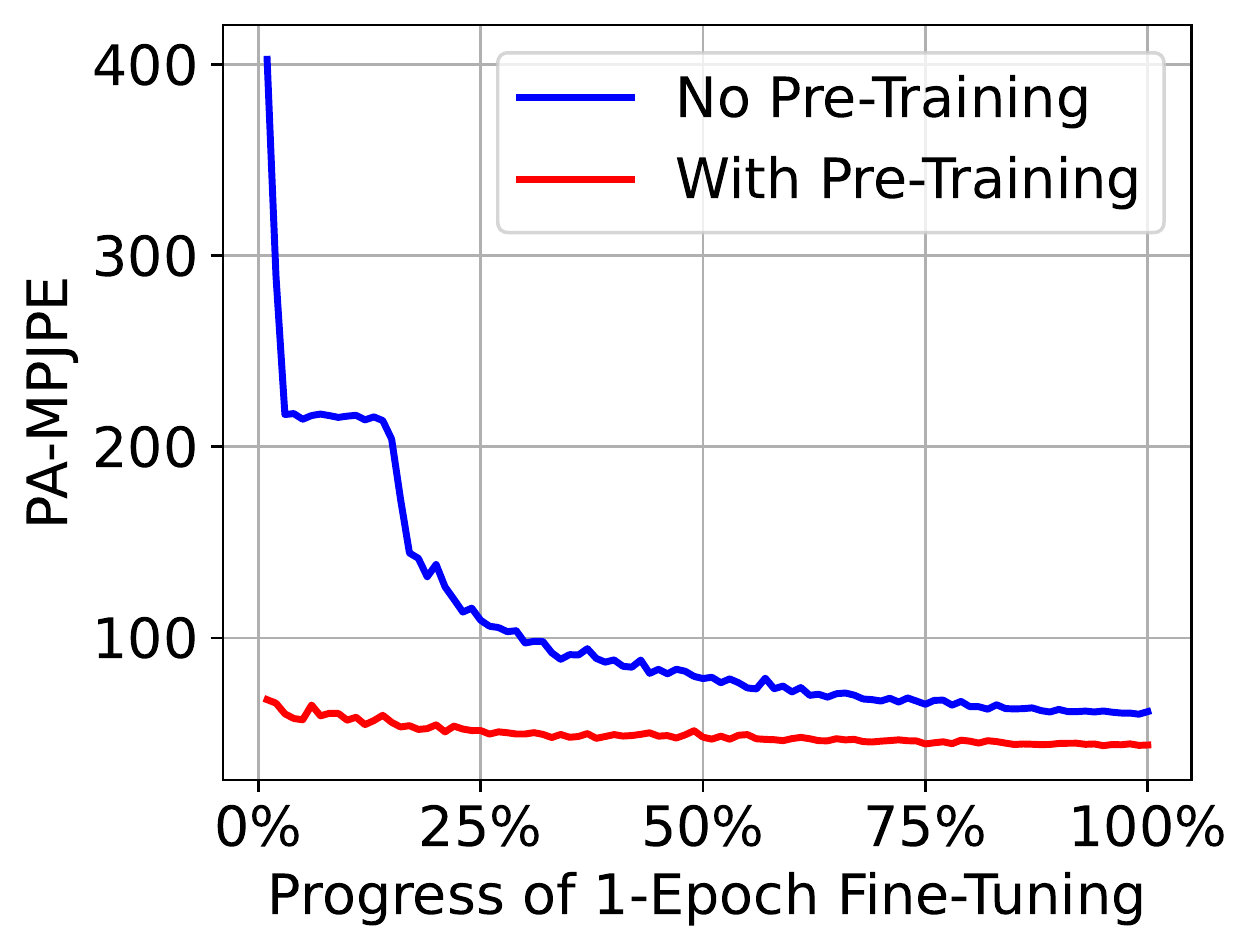}
 }
 \subfloat[(b)][Fine-Tune with Less Data\label{fig:fewshot}]{
   \includegraphics[trim=0 0 0 0, clip,width=0.48\linewidth]{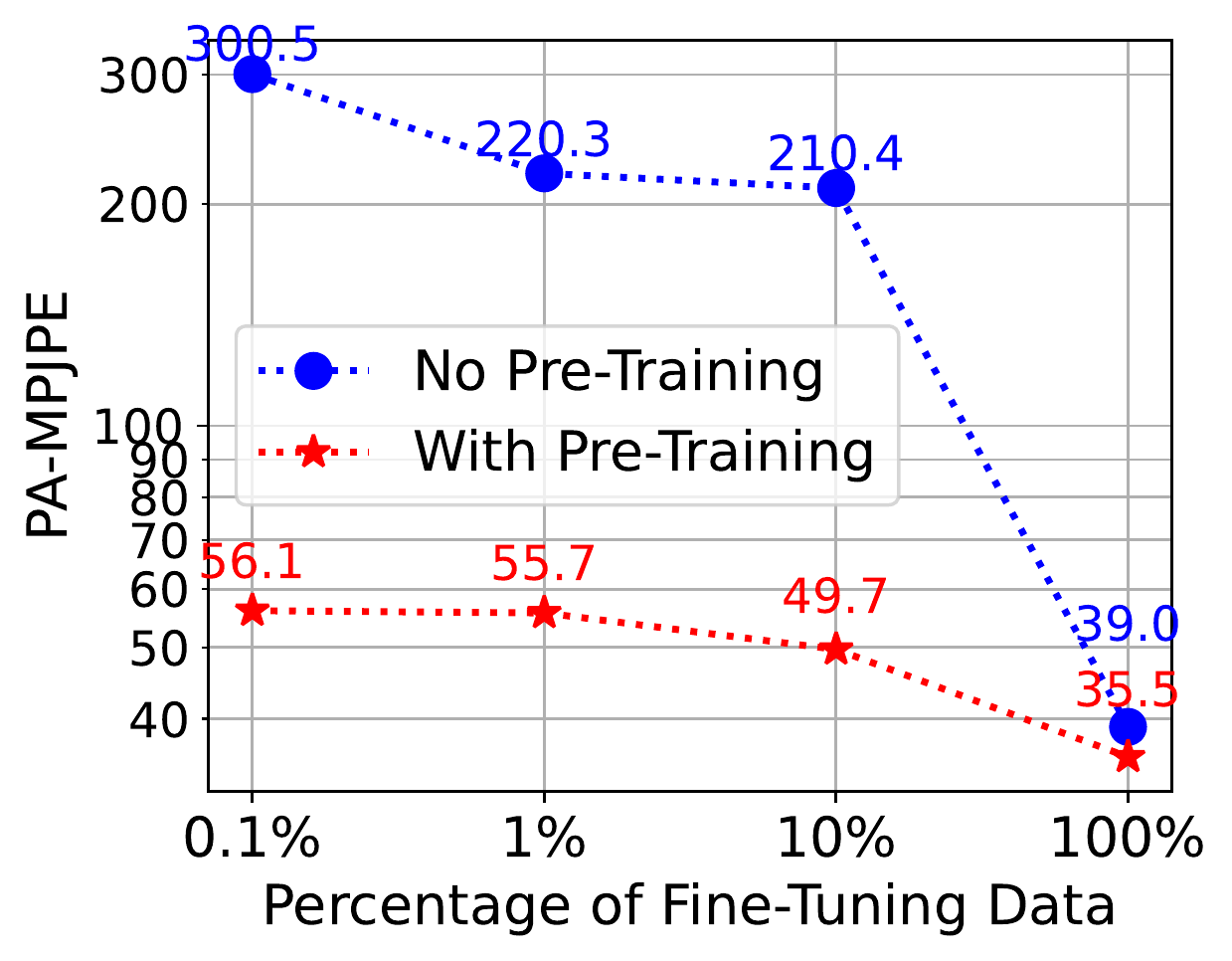}
 }
\caption{
Fine-tuning behavior on Human3.6M. \textbf{(a)} We conduct a 1-epoch fine-tuning and report PA-MPJPE for each fine-tuning step. \textbf{(b)} We use a percentage of Human3.6M data for fine-tuning and report PA-MPJPE. }
\vspace{-4mm}
\end{figure}

\begin{figure*}[t]
\begin{center}
	\centering
\includegraphics[trim=0 0 0 0, clip,width=0.495\textwidth]{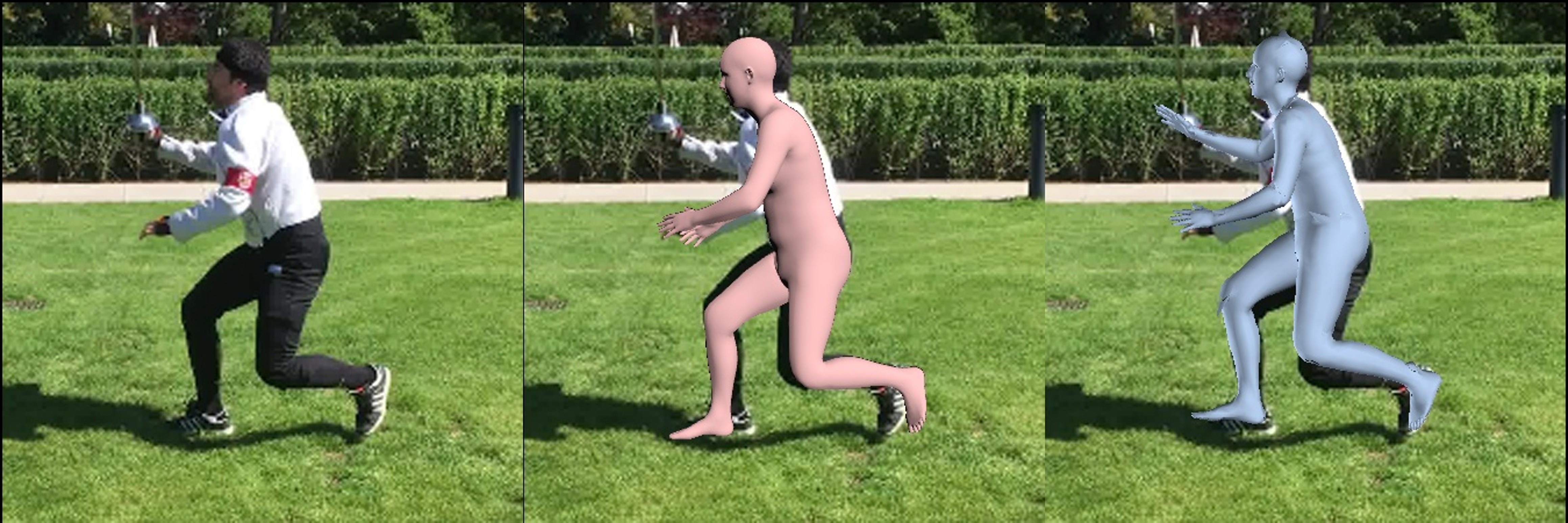}
\includegraphics[trim=0 0 0 0, clip,width=0.495\textwidth]{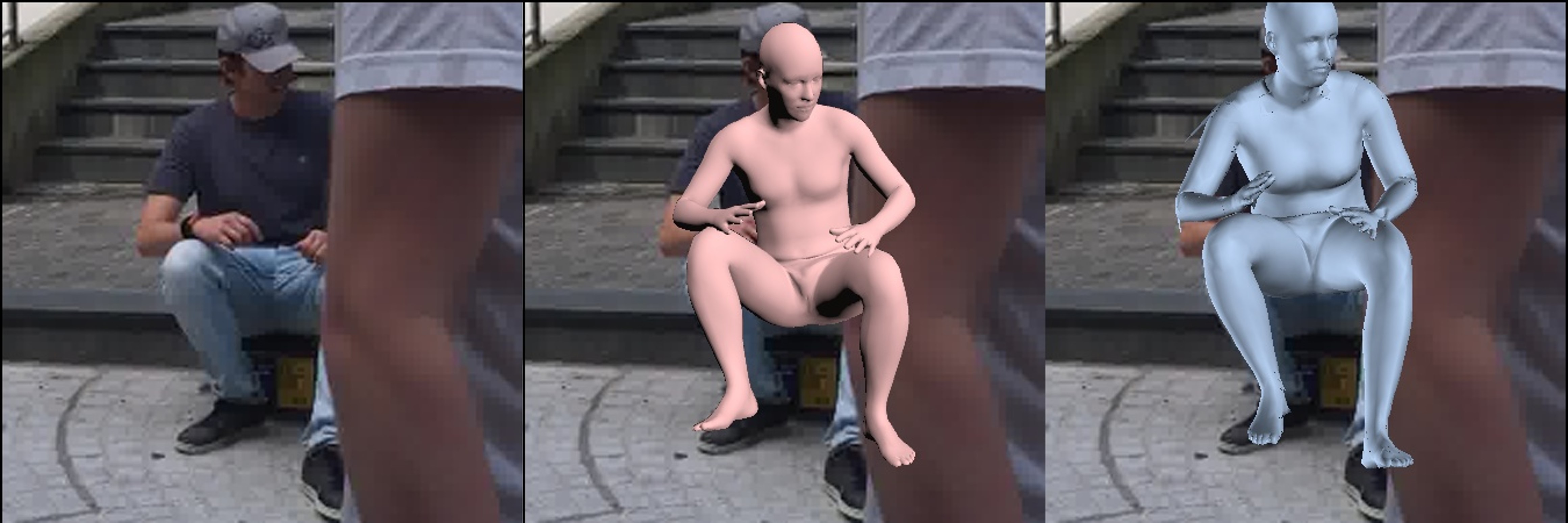}\\
\includegraphics[trim=0 0 0 0, clip,width=0.495\textwidth]{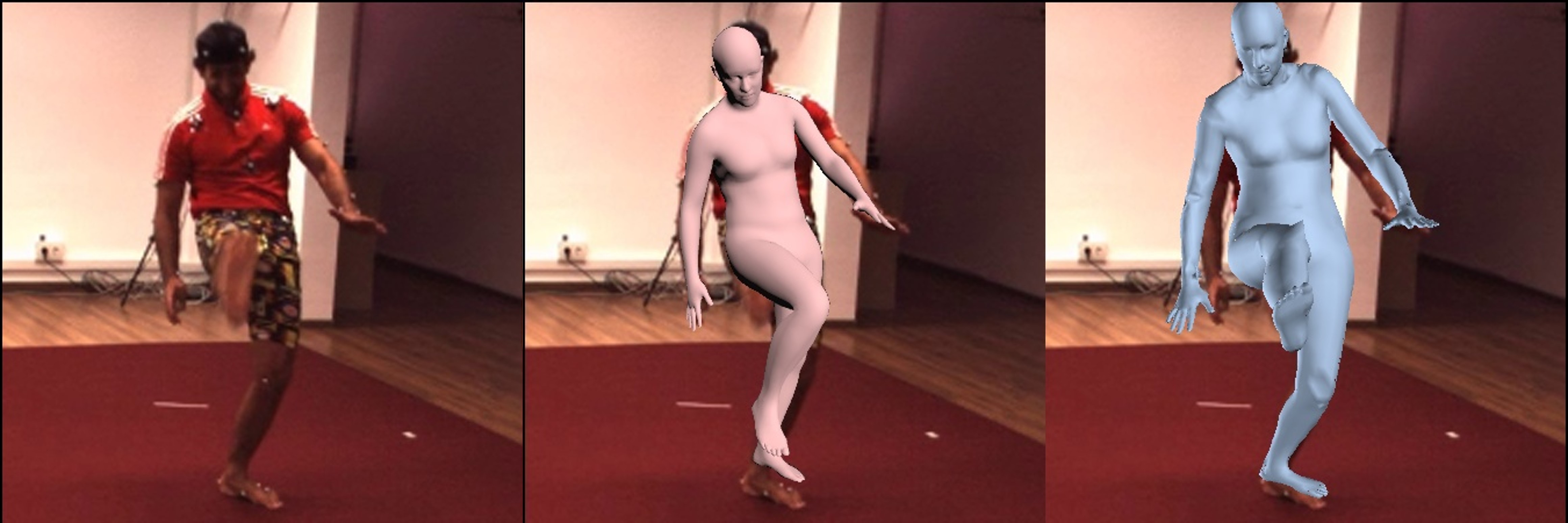}
\includegraphics[trim=0 0 0 0, clip,width=0.495\textwidth]{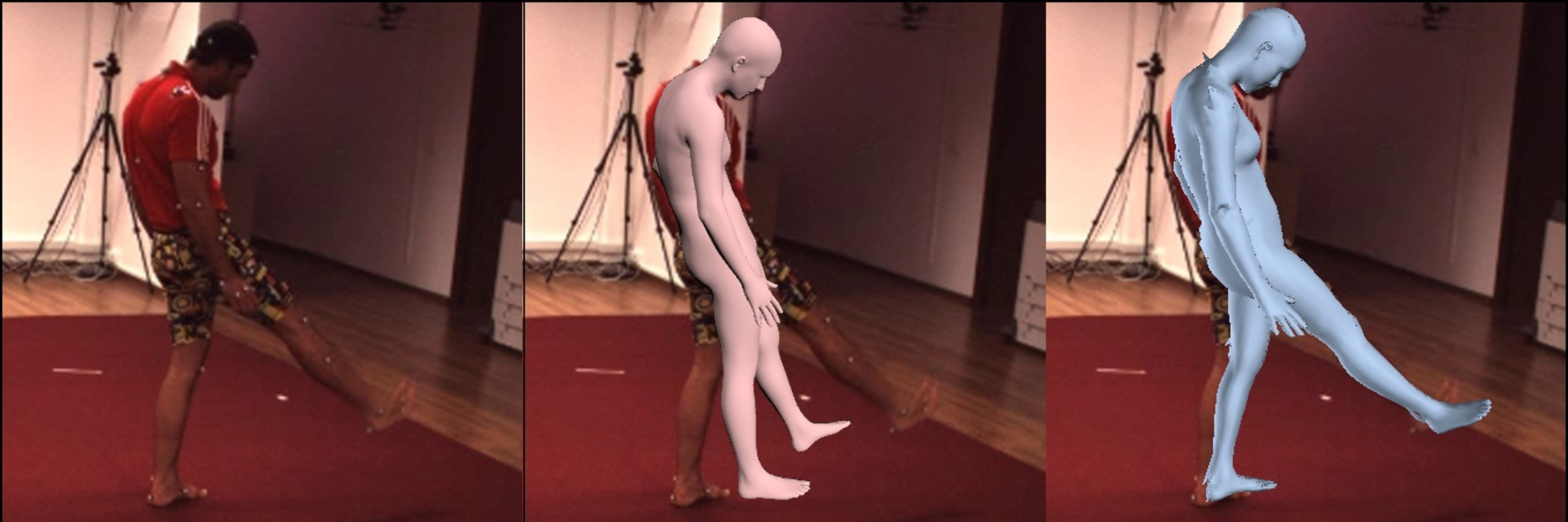}
\resizebox{1.\textwidth}{!}{
\setlength{\tabcolsep}{25pt}
\begin{tabular}{cccccc}
\footnotesize{Input} & \footnotesize{CLIFF} &  \footnotesize{Ours}  &  \footnotesize{Input} &  \footnotesize{CLIFF} & \footnotesize{Ours}
\end{tabular}}
\vspace{-4mm}
\captionof{figure}{
	Qualitative comparison. For each example, we show the results from CLIFF~\cite{li2022cliff} and our proposed MPT. Both CLIFF and MPT generate good quality human meshes, but MPT has more favorable body pose. Tow row: 3DPW. Bottom row: Human3.6M.}
	\label{fig:qualitative-compare}
\end{center}%
\vspace{-6mm}
\end{figure*}

\begin{figure*}
\begin{center}
\rotatebox{90}{ \ \ \ \ \ \ Freeze} \includegraphics[trim=0 0 0 0, clip,width=0.97\textwidth]{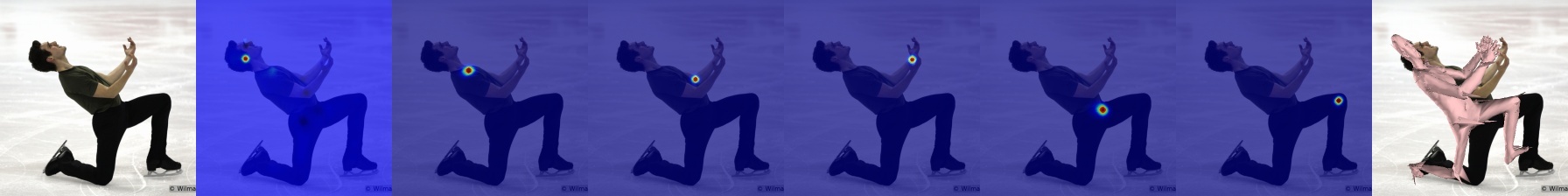}
\rotatebox{90}{ \ \ \ \ FT (Ours)} \includegraphics[trim=0 0 0 0, clip,width=0.97\textwidth]{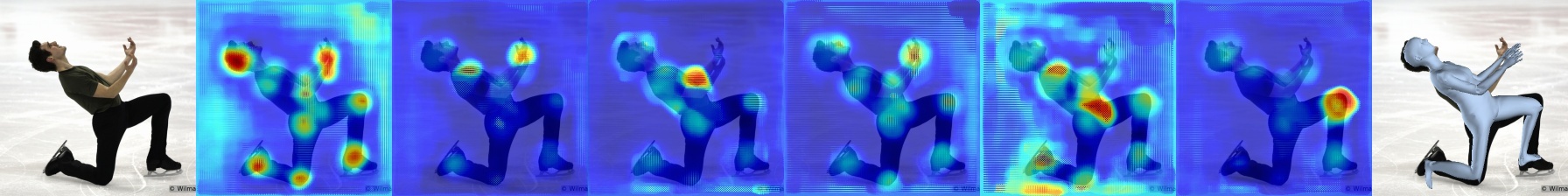}
\rotatebox{90}{ \ \ \ }\resizebox{1.\textwidth}{!}{
\setlength{\tabcolsep}{20pt}
\begin{tabular}{ccccccccc}
 \ \ Input & \ \ L-Ear & \ \ L-Sho & \ \ L-Elb & \ \ L-Wri & \ \ L-Hip & L-Ank & Output
\end{tabular}}
\vspace{-4mm}
\caption{
\revise{Visualization of pose feature maps. In the top row, we freeze the backbone and fine-tune only the mesh regression transformer. However, we observe that the pose feature maps behave like conventional heatmaps, detecting one joint per map and resulting in incorrect reconstruction. In the bottom row, we unfreeze the backbone and perform end-to-end fine-tuning. Our model learns to extract pose feature maps where each map captures information on multiple body joints, leading to improved reconstruction and more accurate pose estimation. 
More visualizations are provided in the supplementary material.}
} 
\label{fig:heatmap}
\vspace{-12mm}
\end{center}
\end{figure*} 

\begin{table*}[t!]\centering
\subfloat[Comparison of Different Pre-Training: MoCap-generated Heatmaps vs. 2D Coordinates 
\label{tbl:pose2mesh}]{\tablestyle{1pt}{1.1}
\begin{tabular}{lcc}
    \toprule[1.5pt]
	Pre-Training & MPJPE $\downarrow$ & PA-MPJPE $\downarrow$ \\ 
	\midrule
	2D Coordinates  & 139.3 & 88.9\\
	MoCap-gen. Heatmaps &  88.9 & 58.4\\
	\bottomrule[1.5pt]
\end{tabular}
}
\hfill
\subfloat[Different Inputs During Fine-Tuning \label{tbl:module-pretrain}]{\tablestyle{2pt}{1.1}
\begin{tabular}{lcccc}
\multicolumn{1}{c}{} \\ 
    \toprule[1.5pt]
	Fine-Tuning & MPT & FT & MPJPE $\downarrow$ & PA-MPJPE $\downarrow$ \\
	\midrule
	Image Features & \xmark & \cmark & 78.0 & 47.0\\
	Pose FeatMaps & \xmark & \cmark & 58.7 & 39.7\\
	Pose FeatMaps & \cmark & \cmark & 53.3 & 35.5\\
	\bottomrule[1.5pt]
\end{tabular}
}
\hfill
\subfloat[Masked Heatmap Modeling 
\label{tbl:mjm}]{\tablestyle{4pt}{1.1}
\begin{tabular}{ccc}
    \multicolumn{1}{c}{} \\ 
    \toprule[1.5pt]
	Method & MPJPE $\downarrow$ & PA-MPJPE $\downarrow$ \\
	\midrule
	MPT w/o MHM & 97.7 & 64.4\\
	MPT w/ MHM & 88.9 & 58.4\\
	\bottomrule[1.5pt]
	\multicolumn{1}{c}{} \\ 
\end{tabular}

}
\caption{
\textbf{(a)} We study pre-training using different input representations to the mesh regression transformer. We evaluate pre-training performance (\textit{i.e.,} testing without fine-tuning) on Human3.6M. \textbf{(b)} During fine-tuning, we study the effect of different input representations to the mesh regression transformer, including image features and pose feature maps. We conduct fine-tuning and evaluation on Human3.6M. \textbf{(c)} Ablation study of Masked Heatmap Modeling. We evaluate our pre-trained MPT on Human3.6M without any fine-tuning.
\label{tab:ablation_heatmaps}
}
\vspace{-4mm}
\end{table*}

\begin{table}[t]
\tablestyle{2pt}{1.1} 
\centering
\begin{tabular}{lccc}
    \toprule[1.5pt]
	Backbone & FT & MPJPE $\downarrow$ & PA-MPJPE $\downarrow$ \\
 	\midrule
	Freeze & \cmark & 78.1 & 53.5\\
        Unfreeze & \cmark & 53.3 & 35.5\\
	\bottomrule[1.5pt]
\end{tabular}
\caption{Comparison between freeze and unfreeze backbone during fine-tuning. We conducted fine-tuning on Human3.6M.
}
\label{tbl:ft_before_after}
\vspace{-5mm}
\end{table}
\Paragraph{Learning Convergence:}
We study the impact of MPT during fine-tuning. We perform a 1-epoch fine-tuning on Human3.6M, and report results for each fine-tuning step. Figure~\ref{fig:converge} shows that, with our pre-training, the fine-tuning converges better than that without pre-training.

\Paragraph{Fine-Tuning with Less Data:} We select 0.1\%, 1\%, and 10\% of Human3.6M training data for fine-tuning, respectively. Figure~\ref{fig:fewshot} shows that our pre-training helps improve the learning performance when less data is used during fine-tuning. For example, our pre-trained model with 0.1\% fine-tune data achieves 56.1 PA-MPJPE, which is much better than 210.4 PA-MPJPE of our non-pretrained model with 10\% fine-tune data.

\Paragraph{Pre-Training with MoCap-generated Heatmaps:} Since we use MoCap-generated heatmaps during our pre-training, one important question is that what if we directly input 2D joint coordinates to the mesh regression transformer. Table~\ref{tbl:pose2mesh} shows the comparison of the pre-training performance (\textit{i.e.,} testing without fine-tuning) on Human3.6M. We observe that the use of MoCap-generated heatmaps effectively improves the pre-training performance.

\Paragraph{Fine-Tuning with Pose Feature Maps:} \revise{During the fine-tuning stage, our mesh regression transformer takes pose feature maps as inputs. One may wonder what if we use image feature maps instead, as discussed in Mesh Graphormer~\cite{lin2021mesh}. To investigate the impact of this design choice, we conducted an ablation study on Human3.6M, as shown in Table~\ref{tbl:module-pretrain}. The results indicate that using pose feature maps during the fine-tuning stage leads to better performance across all considered metrics.}

\Paragraph{Analysis of Pose Feature Maps:} \revise{
In Figure~\ref{fig:heatmap}, we present a visualization of the pose feature maps during fine-tuning. The top row of Figure~\ref{fig:heatmap} shows that when the backbone is frozen during fine-tuning, the pose feature maps behave like traditional heatmaps, capturing a single body joint location per map. We can see that the reconstructed mesh is not correct due to the side view angle and self occlusions. In contrast, as shown in the bottom row of Figure~\ref{fig:heatmap}, when we fine-tune both the backbone and the mesh regression transformer, our model learns to extract pose feature maps where each map captures information on multiple body joints, resulting in improved reconstruction and more accurate pose estimation. For example, in order to predict the location of the left ear, our model pays attention to not only the keypoints on the face, but also the body joints on the arms and legs. In Table~\ref{tbl:ft_before_after}, we provide a quantitative comparison between freezing and unfreezing backbone during fine-tuning. The results suggest end-to-end fine-tuning improves the reconstruction performance.}

\begin{figure*}
\begin{center}
\includegraphics[trim=0 0 0 0, clip,width=0.195\textwidth]{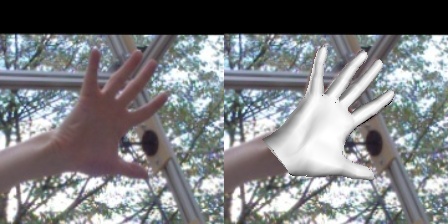}
\includegraphics[trim=0 0 0 0, clip,width=0.195\textwidth]{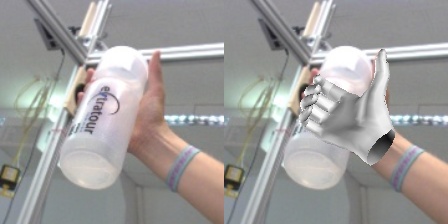}
\includegraphics[trim=0 0 0 0, clip,width=0.195\textwidth]{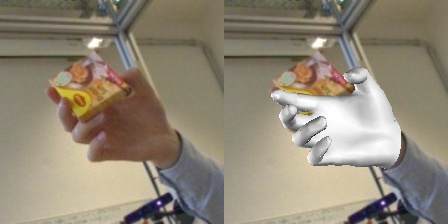}
\includegraphics[trim=0 0 0 0, clip,width=0.195\textwidth]{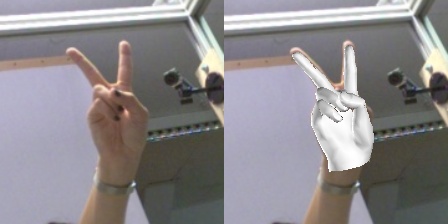}
\includegraphics[trim=0 0 0 0, clip,width=0.195\textwidth]{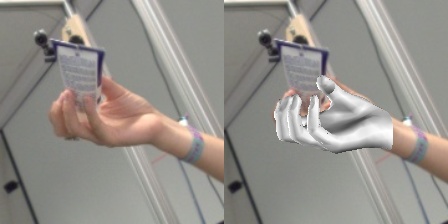}\\
\includegraphics[trim=0 0 0 0, clip,width=0.195\textwidth]{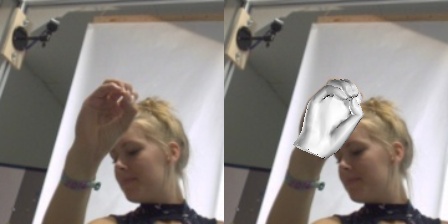}
\includegraphics[trim=0 0 0 0, clip,width=0.195\textwidth]{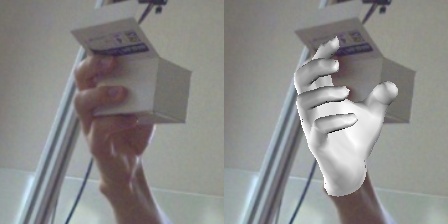}
\includegraphics[trim=0 0 0 0, clip,width=0.195\textwidth]{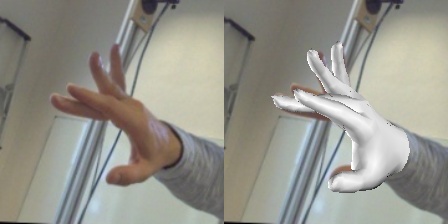}
\includegraphics[trim=0 0 0 0, clip,width=0.195\textwidth]{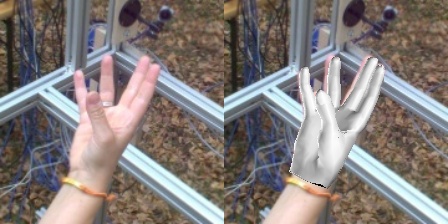}
\includegraphics[trim=0 0 0 0, clip,width=0.195\textwidth]{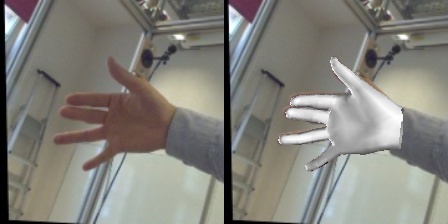}\\
\includegraphics[trim=0 0 0 0, clip,width=0.195\textwidth]{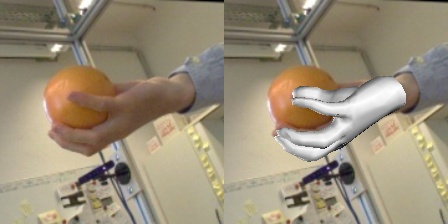}
\includegraphics[trim=0 0 0 0, clip,width=0.195\textwidth]{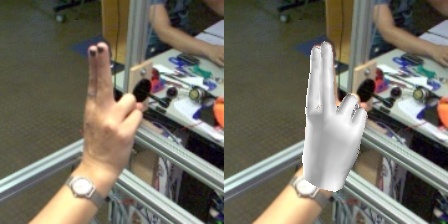}
\includegraphics[trim=0 0 0 0, clip,width=0.195\textwidth]{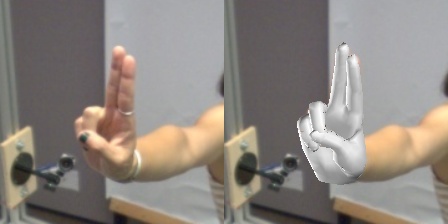}
\includegraphics[trim=0 0 0 0, clip,width=0.195\textwidth]{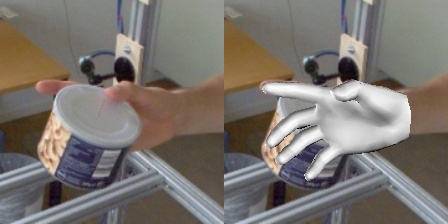}
\includegraphics[trim=0 0 0 0, clip,width=0.195\textwidth]{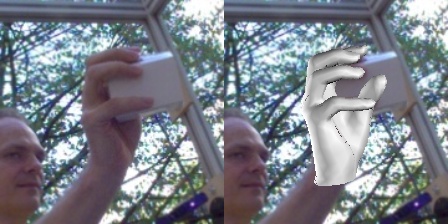}
\caption{
Qualitative results of MPT applying to 3D hand reconstruction on FreiHAND dataset.  
} 
\label{fig:hand}
\vspace{-6mm}
\end{center}
\end{figure*}

\begin{table}
\tablestyle{8pt}{1.1} 
\centering
\resizebox{1.\columnwidth}{!}{
    \begin{tabular}{lccccc}
	\toprule[1.5pt]
	 Percentage of PT Data & $20\%$ & $40\%$ & $60\%$ & $80\%$ & $100\%$\\
	 \midrule
	 PA-MPJPE $\downarrow$ & 63.4 & 62.4 & 59.9 & 59.4 & 58.4\\
	\bottomrule[1.5pt]
	\end{tabular}
}	
\caption{Analysis of pre-training data size. We conduct pre-training using different percentage of data, and evaluate the pre-trained MPT on Human3.6M without fine-tuning.}
\label{tbl:num_pt_data}
\vspace{-4mm}
\end{table}
\begin{table}[t]
\tablestyle{0.8pt}{1.1} 
\centering
\begin{tabular}{lcccc}
    \toprule[1.5pt]
	Method  & PA-MPVPE $\downarrow$ & PA-MPJPE $\downarrow$ & F@5 mm $\uparrow$ & F@15 mm $\uparrow$\\
	\midrule
	Hasson et al~\cite{hasson2019learning} & 13.2 & $-$ & 0.436 & 0.908\\
	Boukhayma et al.~\cite{boukhayma20193d} & 13.0 & $-$ & 0.435 & 0.898\\
	FreiHAND~\cite{zimmermann2019freihand} & 10.7 & $-$ & 0.529 & 0.935\\
	Pose2Mesh~\cite{Choi_2020_ECCV_Pose2Mesh} & 7.8 & 7.7 & 0.674 & 0.969\\
	I2LMeshNet~\cite{Moon_2020_ECCV_I2L-MeshNet} & 7.6 & 7.4 & 0.681 & 0.973\\
    METRO~\cite{lin2020end}  & 6.8 & 6.7 & 0.717 & 0.981\\
    Tang \textit{et al.}~\cite{tang2021towards}  & 6.7 & 6.7 & 0.724 & 0.981\\
    FastMETRO~\cite{cho2022FastMETRO} & $-$ & 6.5 & $-$ & 0.982\\
    Graphormer~\cite{lin2021mesh}   & 6.0 & 5.9 & 0.764 & 0.986\\
    MobRecon~\cite{MobRecon} & 5.7 & 5.8 & 0.784 & 0.986\\
    \midrule
    MPT (Ours) & $\textbf{5.4}$ & $\textbf{5.6}$ & $\textbf{0.789}$ & $\textbf{0.988}$\\
	\bottomrule[1.5pt]
\end{tabular}
\caption{Performance comparison with the previous state-of-the-art methods on FreiHAND dataset. }
\label{tbl:compare-hand}
\vspace{-4mm}
\end{table}

\Paragraph{Effectiveness of Masked Heatmap Modeling:} Table~\ref{tbl:mjm} shows an ablation study of the proposed Masked Heatmap Modeling (MHM), tested on Human3.6M. We observe MHM improves the reconstruction performance by a large margin, especially on PA-MPJPE metric.

\Paragraph{Pre-Training Data Size:}
Since we use 2 million meshes for pre-training, an interesting question is how much data should we use for pre-training. To answer the question, we conduct pre-training using different percentage of the 2 million meshes. We then evaluate our MPT model directly on Human3.6M validation set without fine-tuning. Table~\ref{tbl:num_pt_data} shows that increasing the number of meshes for pre-training can improve the performance on Human3.6M. The results suggest that we could scale-up our pre-training data for further performance improvements, and we leave it as future work. 

\Paragraph{Number of Virtual Camera Views:}
We also study the effect of different number of virtual camera views. Table~\ref{tbl:cam_views} shows that adding more camera views can improve the performance. The results suggest increasing the training data diversity is beneficial to our pre-training.

\Paragraph{Backbone Analysis:} As we use HigherHRNet as our backbone to extract feature maps, one may wonder what if we use other backbones. In Table~\ref{tbl:backbones}, we replace HigherHRNet with an early baseline called SimpleBaseline~\cite{xiao2018simple}. Two backbones have quite different mAPs on COCO keypoint dataset. But after adding the backbone to MPT for training, both MPT variants give similar results, which suggests MPT is not very sensitive to the backbone choices.

\Paragraph{Qualitative Results:} Figure~\ref{fig:qualitative-compare} shows the  qualitative results of MPT compared with CLIFF~\cite{li2022cliff} on Human3.6M and 3DPW datasets. While both methods can generate good quality human meshes, MPT has more favorable body poses when there are self-occlusions in the images. 

\begin{table}
\tablestyle{8pt}{1.1} 
\centering
    \begin{tabular}{lccc}
	\toprule[1.5pt]
	 Number of Views & 1 & 2 & 4\\
	 \midrule
	 PA-MPJPE $\downarrow$ & 61.7 & 61.4 & 58.4 \\
	\bottomrule[1.5pt]
	\end{tabular}
\caption{Ablation study of numbers of virtual camera views. We evaluate the pre-trained MPT on Human3.6M without fine-tuning.}
\vspace{-0mm}
\label{tbl:cam_views}
\end{table}
\begin{table}[t]
\tablestyle{2pt}{1.1} 
\centering
\begin{tabular}{lccc}
    \toprule[1.5pt]
	Backbone & & MPJPE $\downarrow$ & PA-MPJPE $\downarrow$ \\
 	\midrule
	SimpleBaseline~\cite{xiao2018simple} &  & 45.8 & 32.8\\
	HigherHRNet~\cite{cheng2020higherhrnet} & & 46.6 & 32.4\\
	\bottomrule[1.5pt]
\end{tabular}
\caption{Comparison of different human pose networks. We conduct end-to-end fine-tuning on mixed 2D/3D datasets, and evaluate on Human3.6M. No pre-training in this ablation study. 
}
\label{tbl:backbones}
\vspace{-2mm}
\end{table}

\Paragraph{Applying to 3D Hand Reconstruction:} MPT is a generic pre-training scheme for human mesh regression task. We demonstrate the flexibility of MPT on 3D hand reconstruction, and we conduct the experiments on FreiHAND~\cite{zimmermann2019freihand} dataset. Since there is less MoCap data for 3D hands, we use a recently proposed synthetic dataset called Complement~\cite{MobRecon} for pre-training. Table~\ref{tbl:compare-hand} shows the performance comparison with previous works. MPT outperforms the previous state-of-the-art methods, including MobRecon~\cite{MobRecon} which also uses Complement as training data. Figure~\ref{fig:hand} shows the qualitative examples of our hand reconstruction.

\section{Conclusion}\label{sec:conclusion}
We introduced Mesh Pre-Training (MPT), a simple yet effective pre-training strategy that leverages large-scale MoCap data to pretrain the mesh regression transformer for 3D human pose and mesh reconstruction from a single image. We introduce the use of MoCap-generated heatmaps with Masked Heatmap Modeling for pre-training the mesh regression transformer. Experimental results show that our method advances the state-of-the-art performance on Human3.6M, 3DPW, and FreiHAND datasets. We further show that MPT can be directly applied to real images without fine-tuning on any image-mesh pairs.

{\small
\bibliographystyle{ieee_fullname}
\bibliography{humanmesh, vlp}
}

\clearpage
\appendix
\pdfoutput=1
\twocolumn[{%
\renewcommand\twocolumn[1][]{#1}%
\begin{center}
\textbf{\Large Supplementary Material}
\end{center}
\vspace{2mm}
}]

\maketitle
\ificcvfinal\thispagestyle{empty}\fi

\appendix

\begin{figure}[]
\begin{center}
\includegraphics[trim=0 0 0 0, clip,width=0.9\columnwidth]{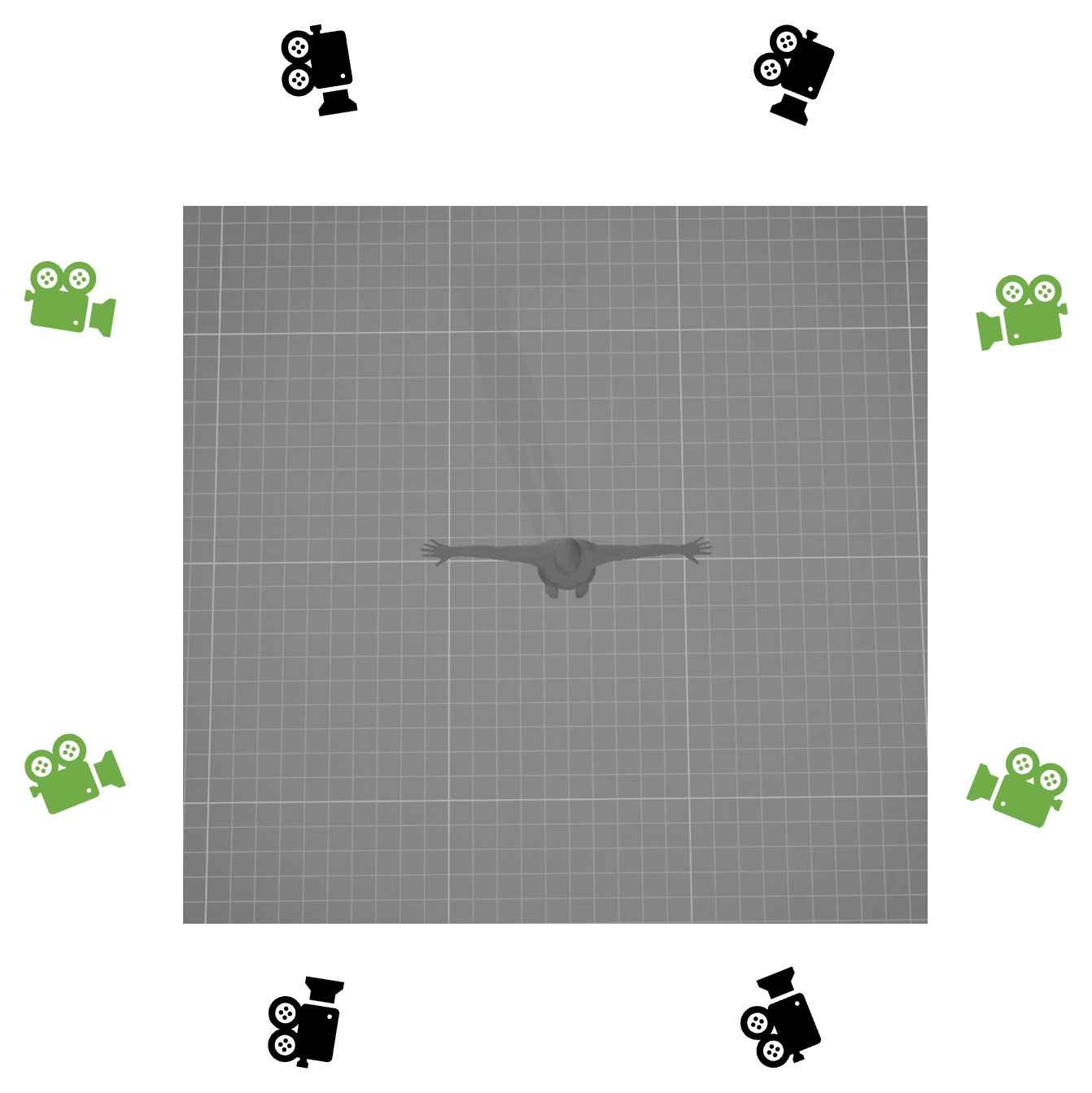}
\caption{
Illustration of our virtual cameras in a top-down view. In order to set up a virtual camera at a proper position, we follow the camera settings of Human3.6M to set up 4 virtual cameras (indicated in black color). In addition, we perform interpolation to obtain extra virtual camera positions (indicated in green color). 
} 
\vspace{-6mm}
\label{fig:virtualcam}
\end{center}
\end{figure}

\section{Details of Virtual Cameras}
During pre-training, we use virtual cameras to help synthesize heatmaps. In order to set up the virtual cameras at proper positions, we follow the camera settings of Human3.6M~\cite{ionescu2014human3} in our experiments. To be specific, we set up 4 virtual cameras using the camera parameters from Subject 1 of Human3.6M training set.

\section{Adding More Virtual Cameras}

\begin{table}
\tablestyle{8pt}{1.1} 
\centering
    \begin{tabular}{ccc}
	\toprule[1.5pt]
	 Number of Views & MPJPE $\downarrow$ & PA-MPJPE $\downarrow$\\
	 \midrule
	 1 & 92.5 & 61.7\\
	 2 & 89.5 & 61.4\\
	 4 & 89.0 & 58.4\\
	 8 & 88.3 & 58.2\\
	\bottomrule[1.5pt]
	\end{tabular}
\caption{Adding more virtual camera views. We evaluate the pre-trained model on Human3.6M validation set without fine-tuning.}
\vspace{-4mm}
\label{tbl:more_cam_views}
\end{table}

We further investigate whether adding more camera views can improve the performance. Given the 4 virtual cameras defined by Human3.6M, we perform interpolation to obtain additional 4 virtual camera views. Figure~\ref{fig:virtualcam} illustrates our virtual cameras using an example. The 4 virtual cameras defined by Human3.6M are denoted in black color. The interpolated virtual cameras are denoted in green color.

In Table~\ref{tbl:more_cam_views}, we observe that adding more camera views in pre-training improves the performance for both metrics on Human3.6M. Note that we evaluate the pre-trained model without fine-tuning.

\section{Visualization of Pose Feature Maps}
\begin{figure*}
\begin{center}
\rotatebox{90}{ \ \ \ \ \ \ Freeze} \includegraphics[trim=0 0 0 0, clip,width=0.97\textwidth]{figs/heatmaps_freeze/10-Joshua-Farris_meshpt_pred.jpg.all.jpg_sampled.jpg}
\rotatebox{90}{ \ \ \ \ FT (Ours)} \includegraphics[trim=0 0 0 0, clip,width=0.97\textwidth]{figs/heatmaps_ours/10-Joshua-Farris_meshpt_pred.jpg.all.jpg_sampled.jpg}
\rotatebox{90}{ \ \ \ \ \ \ Freeze} \includegraphics[trim=0 0 0 0, clip,width=0.97\textwidth]{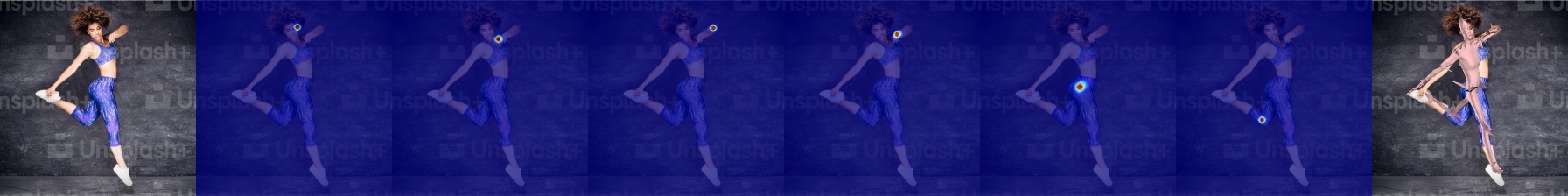}
\rotatebox{90}{ \ \ \ \ FT (Ours)} \includegraphics[trim=0 0 0 0, clip,width=0.97\textwidth]{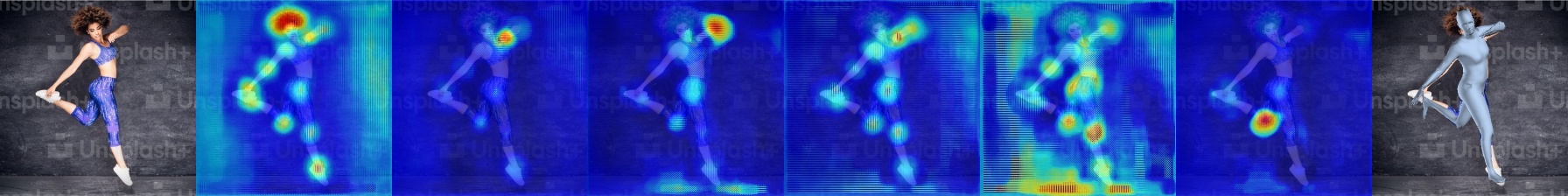}
\rotatebox{90}{ \ \ \ \ \ \ Freeze} \includegraphics[trim=0 0 0 0, clip,width=0.97\textwidth]{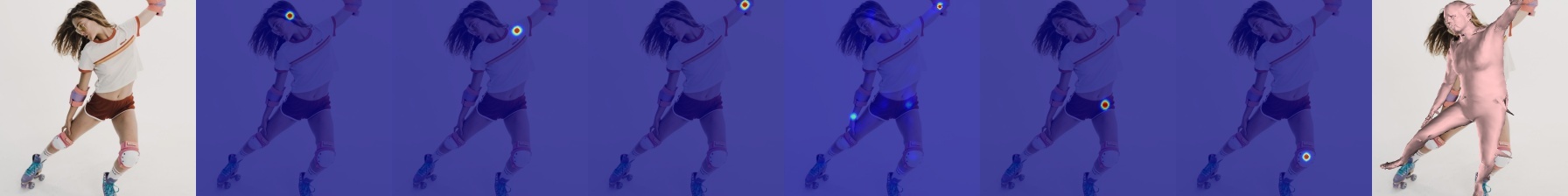}
\rotatebox{90}{ \ \ \ \ FT (Ours)} \includegraphics[trim=0 0 0 0, clip,width=0.97\textwidth]{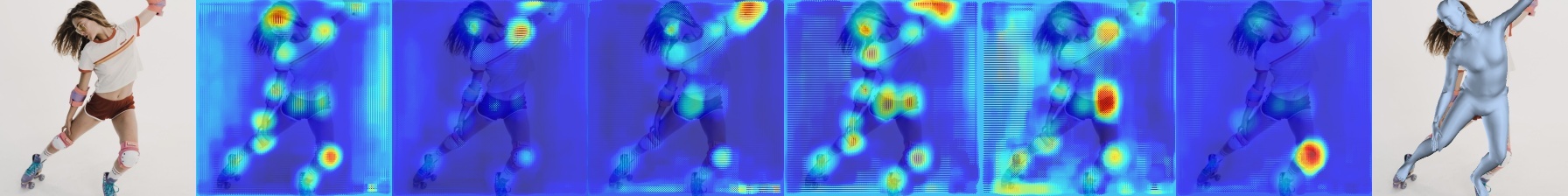}
\rotatebox{90}{ \ \ \ \ \ \ Freeze} \includegraphics[trim=0 0 0 0, clip,width=0.97\textwidth]{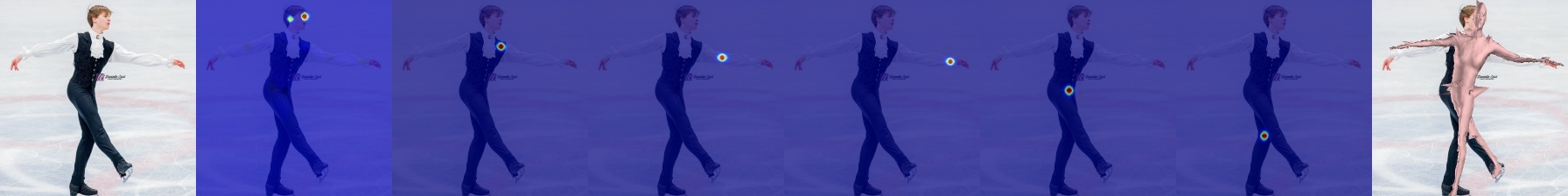}
\rotatebox{90}{ \ \ \ \ FT (Ours)} \includegraphics[trim=0 0 0 0, clip,width=0.97\textwidth]{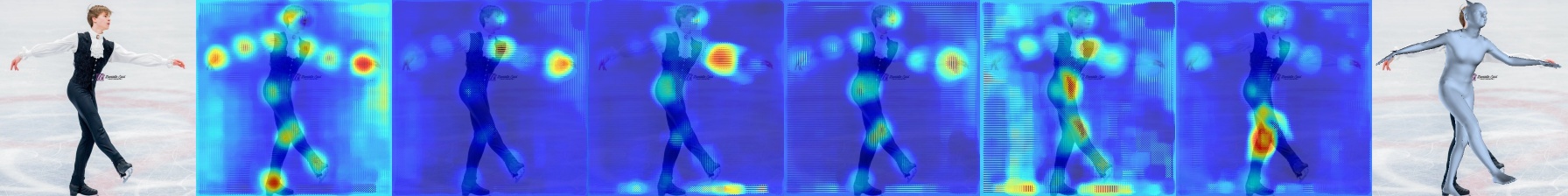}
\rotatebox{90}{ \ \ \ }\resizebox{1.\textwidth}{!}{
\setlength{\tabcolsep}{20pt}
\begin{tabular}{ccccccccc}
 \ \ Input & \ \ L-Ear & \ \ L-Sho & \ \ L-Elb & \ \ L-Wri & \ \ L-Hip & L-Ank & Output
\end{tabular}}
\vspace{-4mm}
\caption{
\revise{Visualization of pose feature maps. Our model learns to extract pose feature maps where each map captures information on multiple body joints, leading to improved reconstruction.}
} 
\label{fig:heatmap_supp}
\vspace{-8mm}
\end{center}
\end{figure*}

In Figure~\ref{fig:heatmap_supp}, we present additional visualization of the pose feature maps. By fine-tuning our pre-trained MPT model, we empirically observed that our model learns to extract pose feature maps where each map captures information on multiple body joints, and our model generates a human mesh with more reasonable pose and shape.

\section{Resolution of Pose Feature Maps}
\begin{table}[t]
\tablestyle{8pt}{1.1} 
\centering
\begin{tabular}{ccc}
    \toprule[1.5pt]
	Resolution & MPJPE $\downarrow$ & PA-MPJPE $\downarrow$ \\
	\midrule
	$112 \times 112$ & 46.5 & 32.7\\
	$224 \times 224$ & 45.3 & 31.7\\
	\bottomrule[1.5pt]
\end{tabular}
\caption{Different resolutions of the pose feature maps. We conduct fine-tuning on a mixture of 2D and 3D training sets, and evaluate the performance on Human3.6M validation set.}
\label{tbl:resolusion}
\vspace{-2mm}
\end{table}
As we use HigherHRNet~\cite{cheng2020higherhrnet} to generate pose feature maps, one may wonder what resolution is needed. Table~\ref{tbl:resolusion} shows the performance comparison with different resolutions. The results suggest that using larger resolution (\textit{i.e.,} $224 \times 224$) can slightly improve the results. We note that HigherHRNet was pre-trained using the non-cropped images~\cite{cheng2020higherhrnet}, but we use $224\times224$ cropped images instead. We think that increasing the input resolution to the original resolution on which HighHRNet was trained could further enhance model performance.

\section{Additional Implementation Details}

We implement our models based on PyTorch~\cite{paszke2019pytorch} and Graphormer codebase~\cite{lin2020end,lin2021mesh}. We additionally adopt DeepSpeed~\cite{rasley2020deepspeed} which empirically leads to faster and more stable training. We use 16 NVIDIA V100 GPUs for most of our training experiments. We
use the Adam optimizer for training. We set the initial learning rate as $2 \times 10^{-4}$, and then use learning rate warmup over the first 10$\%$ training steps followed by linear decay to 0. We perform mesh pre-training on the 2 million meshes (sparsely sampled from AMASS dataset~\cite{mahmood2019amass}) for 10 epochs. When fine-tuning on mixed datasets, we empirically fine-tune for 80 epochs.

We use the same transformer architecture as in the literature~\cite{lin2020end,lin2021mesh}. Specifically, our transformer model has 3 transformer blocks. Each block has 4 transformer layers and 4 attention heads. For the 3 transformer blocks, the hidden sizes are 1024, 256, 64, respectively. To reduce the computational cost, the transformer model outputs a coarse mesh. We then use MLPs to upsample the predicted mesh to the original resolution, similar to~\cite{lin2020end,lin2021mesh}. For simplicity, we do not add graph convolutions to the transformer layers. 

Regarding the details of 2D pose estimation model, we use HigherHRNet~\cite{cheng2020higherhrnet} based on {\fontfamily{cmtt}\selectfont HRNet-w48} architecture. In our ablation study, we replace HigherHRNet with SimpleBaseline~\cite{xiao2018simple}, which is based on {\fontfamily{cmtt}\selectfont ResNet101} architecture. Both models are initialized with COCO Keypoint pre-trained weights.

Following the literature~\cite{kolotouros2019convolutional,lin2021mesh}, we use a weak perspective camera model for calculating 2D re-projection loss. Our model predicts camera parameters, including a scaling factor $s$ and a 2D translation vector $t$. Our method does not leverage ground truth camera parameters. The camera parameters are learned by optimizing 2D re-projection.

\section{Limitations and Societal Impact}

3D pose estimation models can be potentially applied to human activity analysis applications, such as detecting whether the senior subjects are falling. However, there is a risk associated with directly applying the models for mission-critical decision making, especially in the health care field. Real-world applications may require an auxiliary supervision model or task-specific fine-tuning.

Our models have a dependency on existing MoCap training data. Some of the public released MoCap data may be licensed, meaning they can only be used for scientific research purposes. Users must strictly adhere to the data usage agreement to utilize MoCap datasets for their intended purposes.

\end{document}